%% file: main.tex
\documentclass[sigconf,nonacm]{acmart}
\def\buildwithsupplement{1}
\newif\ifincludesupplement
\ifdefined\buildwithsupplement
  \includesupplementtrue
\else
  \includesupplementfalse
\fi
\usepackage{bm}
\usepackage{multirow}
\usepackage{colortbl}
\usepackage[most]{tcolorbox}
\usepackage{fontspec}
\usepackage{placeins}
\usepackage{flafter}
\usepackage{stfloats}
\usepackage{afterpage}
\definecolor{mediblue}{HTML}{0039A6}
\definecolor{mediorange}{HTML}{F4511E}
\definecolor{ResultHighlight}{HTML}{F7E6BD}
\definecolor{CNSBlue}{HTML}{3C5488}
\definecolor{CNSCyan}{HTML}{4DBBD5}
\definecolor{CNSTeal}{HTML}{00A087}
\definecolor{CNSWarm}{HTML}{E64B35}
\definecolor{CNSPale}{HTML}{F4F7FA}
\newfontfamily\figurefont{texgyreheros-regular.otf}[
  BoldFont=texgyreheros-bold.otf
]
\providecommand{\gB}{\mathcal{B}}
\providecommand{\gR}{\mathcal{R}}

\providecommand{\gP}{\mathcal{P}}

\AtBeginDocument{%
  }

\setcopyright{none}
\renewcommand\footnotetextcopyrightpermission[1]{}


\begin{document}
\raggedbottom

\title{MediSkill-Evo: Process-Constrained Self-Evolution for Evidence-Grounded Clinical Interaction}

\author{Ruoyu Wu\texorpdfstring{\textsuperscript{1,2,3,*}}{}\quad
Shenfu Xie\texorpdfstring{\textsuperscript{1,2,3,*}}{}\quad
Yinqian Sun\texorpdfstring{\textsuperscript{1,2,3,4}}{}\quad
Haibo Tong\texorpdfstring{\textsuperscript{1,2,3,5}}{}\quad
Feifei Zhao\texorpdfstring{\textsuperscript{1,2,3,5,\textdagger}}{}}
\affiliation{%
  \institution{\textsuperscript{1}Brain-inspired Cognitive AI Lab, Institute of Automation, Chinese Academy of Sciences; \textsuperscript{2}Beijing Key Laboratory of Safe AI and Superalignment; \textsuperscript{3}Beijing Institute of AI Safety and Governance; \textsuperscript{4}School of Artificial Intelligence, University of Chinese Academy of Sciences; \textsuperscript{5}Long-term AI}
  \city{Beijing}
  \country{China}}
\renewcommand{\shortauthors}{Wu et al.}

\begin{abstract}
Interactive clinical agents operate under partial observability, so reliable care depends on reaching the correct diagnosis through evidence-grounded, safe interactions. Yet existing agents struggle to convert experience into reusable process knowledge with explicit provenance and authority. To address this gap, we introduce \textbf{MediSkill-Evo}, which self-evolves governed process knowledge without fine-tuning the backbone. It realizes this self-evolution by updating clinical, process, symbolic, and visual knowledge in four typed banks under type-specific validation and scope rules. The Process-Constrained Preference Harness then turns validated knowledge into action by grounding candidates in evidence and prioritizing safer decisions. We evaluate on 300 MIMIC-IV-derived FullChain encounters, 180 hard-isolation conditions covering six process obligations, and 100 multimodal NEJM image-diagnosis cases. On Qwen FullChain, MediSkill-Evo improves diagnosis accuracy by 7.81\% and treatment-intent coverage by 70.67\% over the best-performing prior agent, while reducing critical failures by 43.04\%. Under stress, it improves the stress-process composite by 7.77\% and required-action completion by 12.41\% over the best-performing agent for each metric, with stronger patient-fact, temporal-evidence, and triage-red-flag recovery and no controller-scored errors in unavailable-evidence, treatment, and triage safety checks. On multimodal NEJM diagnosis, MediSkill-Evo with optional MedSAM localization improves diagnosis accuracy by 2.56\% and core score by 18.96\% over the best-performing memory agent. Code is available at \url{https://anonymous.4open.science/r/mediskill-evo_anonymous-68E7}.
\end{abstract}


\maketitle
\begingroup
\renewcommand{\thefootnote}{*}\footnotetext{Ruoyu Wu and Shenfu Xie contributed equally to this work.}
\renewcommand{\thefootnote}{\textdagger}\footnotetext{Feifei Zhao is the corresponding author.}
\endgroup

\begin{figure*}[!t]
  \centering
  \includegraphics[
    width=0.98\textwidth,
    height=0.36\textheight,
    keepaspectratio
  ]{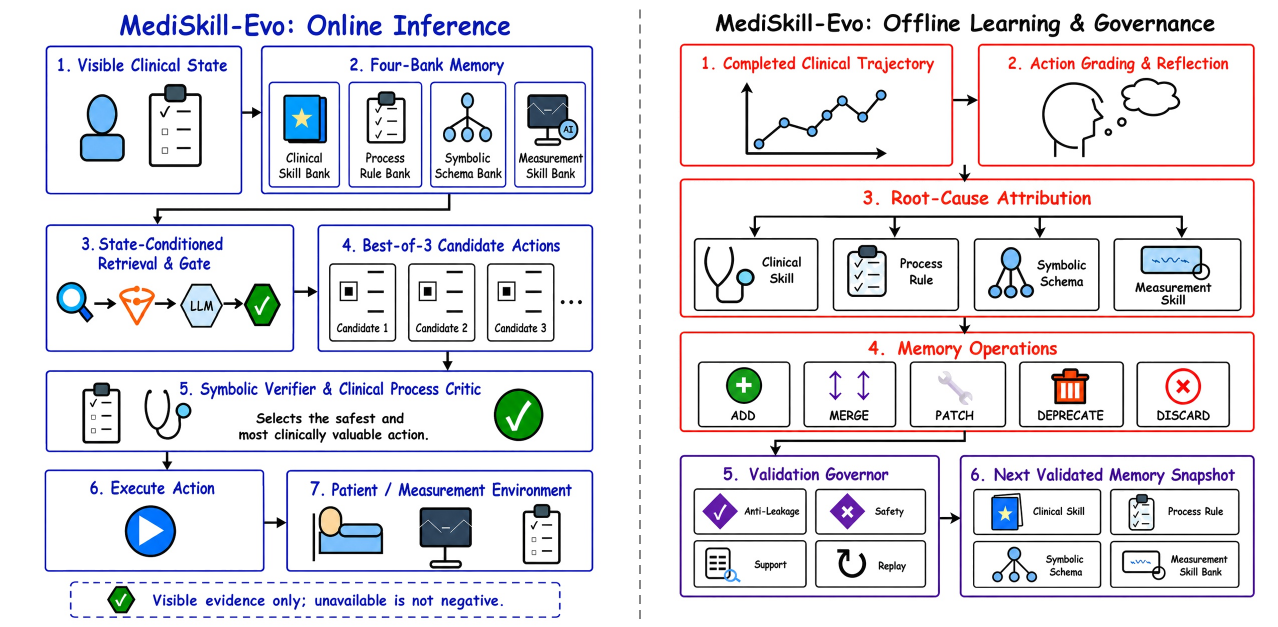}
  \caption{Online inference and offline learning and governance in MediSkill-Evo. Left: the online pipeline retrieves four-bank context, generates and verifies candidate actions, executes the selected action, and incorporates environment feedback. Right: completed trajectories are reflected into typed memory operations, validated for label/evidence leakage, controller-defined safety, support, and replayability, and published as the next memory snapshot.}
  \Description{Two equally sized MediSkill-Evo diagrams separated by a vertical dashed line. The left panel presents the seven-stage online inference pipeline, and the right panel presents the six-stage offline learning and governance pipeline.}
  \label{fig:online-offline-overview}
\end{figure*}

\afterpage{\afterpage{%
\setcounter{dbltopnumber}{2}%
\begin{figure}[!t]
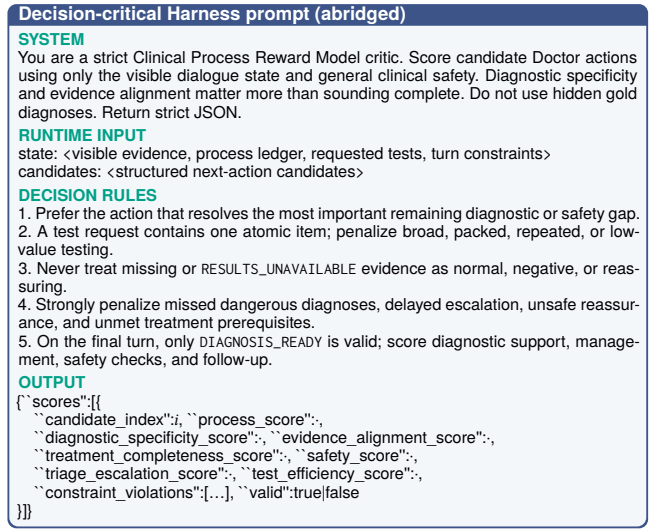

  \centering
  \begin{tcolorbox}[
    width=\columnwidth,
    colback=CNSPale,colframe=CNSBlue,
    colbacktitle=CNSBlue,coltitle=white,
    title={Decision-critical Harness prompt (abridged)},
    fonttitle=\figurefont\bfseries\footnotesize,
    fontupper=\figurefont\fontsize{6.15}{6.85}\selectfont,
    boxrule=0.7pt,arc=1.2mm,outer arc=1.2mm,
    boxsep=0pt,left=1.2mm,right=1.2mm,top=0.9mm,bottom=0.9mm]
\textcolor{CNSTeal}{\textbf{SYSTEM}}\par
You are a strict Clinical Process Reward Model critic. Score candidate Doctor actions using only the visible dialogue state and general clinical safety. Diagnostic specificity and evidence alignment matter more than sounding complete. Do not use hidden gold diagnoses. Return strict JSON.\par\smallskip
\textcolor{CNSTeal}{\textbf{RUNTIME INPUT}}\par
state: \textless visible evidence, process ledger, requested tests, turn constraints\textgreater\par
candidates: \textless structured next-action candidates\textgreater\par\smallskip
\textcolor{CNSTeal}{\textbf{DECISION RULES}}\par
1. Prefer the action that resolves the most important remaining diagnostic or safety gap.\par
2. A test request contains one atomic item; penalize broad, packed, repeated, or low-value testing.\par
3. Never treat missing or \texttt{RESULTS\_UNAVAILABLE} evidence as normal, negative, or reassuring.\par
4. Strongly penalize missed dangerous diagnoses, delayed escalation, unsafe reassurance, and unmet treatment prerequisites.\par
5. On the final turn, only \texttt{DIAGNOSIS\_READY} is valid; score diagnostic support, management, safety checks, and follow-up.\par\smallskip
\textcolor{CNSTeal}{\textbf{OUTPUT}}\par
\{``scores'':[\{\par
\quad ``candidate\_index'':$i$, ``process\_score'':$\cdot$,\par
\quad ``diagnostic\_specificity\_score'':$\cdot$, ``evidence\_alignment\_score'':$\cdot$,\par
\quad ``treatment\_completeness\_score'':$\cdot$, ``safety\_score'':$\cdot$,\par
\quad ``triage\_escalation\_score'':$\cdot$, ``test\_efficiency\_score'':$\cdot$,\par
\quad ``constraint\_violations'':[\ldots], ``valid'':true$|$false\par
\}]\}
  \end{tcolorbox}
  \caption{The decision-critical Clinical Process Critic template. Runtime slots are populated at each turn; retry-only instructions are omitted.}
  \Description{An abridged prompt template with system role, runtime inputs, five decision rules, and a structured candidate-scoring output.}
  \label{fig:harness-core-prompt}
\end{figure}%
}}

\section{Introduction}
Interactive clinical decision-making is the process of gathering evidence over multiple turns and using it to update diagnostic and treatment decisions. Since an agent observes only part of the patient state, it must progressively reduce uncertainty by eliciting decisive history, requesting appropriate examinations, interpreting returned results, and revising its assessment before recommending treatment under safety and urgency constraints. The reliability of this process depends on both the final diagnosis and the path taken to reach it. Unavailable evidence must remain unknown, mandatory care steps must be preserved, and a correct diagnosis cannot justify unsupported or unsafe actions. Past encounters can guide future decisions, but transferred knowledge must retain its provenance and scope and receive only the decision authority warranted by its evidence. We therefore study how a clinical agent can self-evolve from prior trajectories without weakening the evidence boundaries and care obligations that make its actions trustworthy.

Medical agents increasingly support interactive clinical work. AgentClinic~\cite{schmidgall2024agentclinic} models partially observable doctor--patient--measurement encounters; MDAgents~\cite{kim2024mdagents} adapts collaboration to case complexity; EHRAgent~\cite{shi2024ehragent} and MMedAgent~\cite{li2024mmedagent} connect models to executable EHR code and multimodal tools; ReflecTool~\cite{liao2025reflectool} verifies tool use from experience; and MEDDxAgent~\cite{rose2025meddxagent} coordinates specialized modules for interactive differential diagnosis. AI Hospital measures symptom collection, examination choice, and diagnosis in multi-turn simulation~\cite{fan2025aihospital}, while 3MDBench studies multimodal telemedical dialogue~\cite{sviridov2025threem}. Together, they establish interaction, specialization, tool use, and workflow-level evaluation as an emerging baseline rather than a contribution unique to this paper. Our narrower question is how trajectory-derived knowledge with different epistemic roles can be published and exercised through type-dependent validation and decision authority.

Self-evolving agents provide a parameter-efficient route to this goal. Reflexion~\cite{shinn2023reflexion} stores verbal feedback and ExPeL~\cite{zhao2024expel} consolidates cross-trial insights; Voyager~\cite{wang2024voyager}, ICAL~\cite{sarch2024ical}, and Agent Workflow Memory~\cite{wang2025awm} distill executable skills, multimodal abstractions, or workflows. MemP~\cite{fang2025memp} builds updateable procedural instructions, SkillWeaver~\cite{zheng2025skillweaver} discovers reusable skills through practice, and MemBench separates factual and reflective memory while evaluating effectiveness, efficiency, and capacity~\cite{tan2025membench}. These methods show that completed trajectories can become reusable external knowledge without backbone updates. MediSkill-Evo does not claim the first structured memory or workflow evaluation; it proposes a particular complete-system interface in which artifact type controls validation, retrieval scope, and benchmark-time authority. Clinical interaction motivates preventing inferred, missing, or tool-derived information from silently becoming observed fact.

Agent harnesses determine whether typed knowledge actually changes behavior. ReAct~\cite{yao2023react} interleaves reasoning and environment actions; AgentBench~\cite{liu2024agentbench} and AgentBoard~\cite{ma2024agentboard} evaluate multi-step progress; AppWorld~\cite{trivedi2024appworld}, T-Eval~\cite{chen2024teval}, ToolSandbox~\cite{lu2025toolsandbox}, and $\tau$-bench~\cite{yao2025taubench} expose executable state transitions, tool policies, and interaction reliability. In medicine, the harness must additionally bind every result to a valid request, keep unavailable evidence unknown, preserve registered process obligations, and apply benchmark-defined safety checks without access to the hidden diagnosis. The proposed interface turns evolving knowledge into controller-valid, evidence-grounded action; it is not a claim of independent clinical legality or safety.

To govern how experience becomes reusable and how it influences subsequent decisions, we introduce \textbf{MediSkill-Evo}, a clinical agent framework that organizes evolving knowledge by its role in the care process. Completed trajectories propose updates to four typed banks for clinical strategies, process constraints, evidence provenance, and visual procedures. Each bank follows type-specific validation and publication rules before its contents enter the next frozen snapshot. During an encounter, the Process-Constrained Preference Harness retrieves state-relevant knowledge, verifies candidate actions against symbolic and process constraints, and selects among valid alternatives through a safety-prioritized Clinical Process Critic. The central insight is that dependable memory requires knowledge type to determine its validation, retrieval scope, and decision authority.

We evaluate MediSkill-Evo across two backbone endpoints and three complementary settings covering end-to-end FullChain encounters, controlled hard-isolation stress conditions, and multimodal NEJM diagnosis. Across FullChain, MediSkill-Evo consistently improves diagnosis accuracy and treatment-intent coverage while reducing critical failures relative to prior agents. Under stress, its clearest advantages lie in recovering patient facts, temporal evidence, and triage red flags. On multimodal NEJM cases, MediSkill-Evo with optional MedSAM localization also outperforms the strongest memory agent in diagnosis accuracy and core score.

Our contributions are threefold:
\begin{itemize}
  \item We propose MediSkill-Evo, a complete clinical agent system that combines four-bank self-evolution with a Process-Constrained Preference Harness to govern how trajectory-derived knowledge is validated, retrieved, and authorized at decision time.
  \item We introduce FullChain for history-to-prescription interaction, a controlled hard-isolation benchmark for care-process failures, and a request-gated NEJM visual-tool interface for multimodal diagnosis.
  \item Across two backbone endpoints and three complementary evaluation settings, complete-system comparisons report gains in diagnosis, treatment-intent coverage, evidence acquisition, process correctness, and automatically scored safety over prior agents.
\end{itemize}

\section{Method}

\subsection{Overall Architecture}
\label{sec:method-overview}

MediSkill-Evo supports partially observable, interactive clinical decision making while continually revising external knowledge from completed training trajectories rather than updating backbone parameters. At turn $t$, the visible state contains only the presented case information, current observation, dialogue history, and examinations that have been requested and returned; reference diagnoses, evaluator labels, and unrevealed results remain inaccessible. Given the four-bank snapshot $\gB_e$ published after evolution stage $e$, the system retrieves relevant, provenance-traceable knowledge. The Doctor uses the resulting augmented state $s_t$ to ask a history question, request an examination, or submit a clinical plan, and new evidence enters the next state:
\begin{equation}
  v_t=(x^{\mathrm{vis}},o_t,h_t,z_t),\qquad
  u_t=\gR(v_t;\gB_e),\qquad
  s_t=(v_t,u_t).
  \label{eq:augmented-state}
\end{equation}
This formulation separates case-visible evidence from external knowledge and prevents retrieval from introducing hidden labels for the current case.

Figure~\ref{fig:online-offline-overview} summarizes the online inference and offline governance workflows of MediSkill-Evo. MediSkill-Evo consists of a four-bank self-evolution layer and a \emph{Process-Constrained Preference Harness}. After each training case, the former routes trajectory-derived experience to the Clinical Skill, Process Rule, Symbolic Schema, and Measurement banks, which respectively represent clinical strategies, cross-case workflow rules, evidence boundaries, and visual measurement procedures; provenance, safety, and replay checks govern publication of the next frozen snapshot. Within an encounter, the Harness retrieves state-relevant objects from these banks. It directly executes a rule-mandated process checkpoint when one is triggered; otherwise, it generates structured candidates and selects an executable action through symbolic verification and Clinical Process Critic comparison. Here, \emph{preference} denotes test-time ranking of candidates generated for the same state, not parameter-level preference learning or reinforcement learning. Encounter traces support only later evolution stages and never revise a case in progress.

\subsection{Four-Bank Self-Evolution}
\label{sec:four-bank-evolution}

Four-bank self-evolution converts completed case trajectories into reusable, auditable external knowledge. For training case $i$, $\tau_i=(\{v_t,u_t,a_t,o_{t+1}\}_{t=0}^{T},y_i)$ records visible states, retrieved knowledge, executed actions, environment responses, and the post-encounter evaluation, separating what the system observed, retrieved, and executed. Only after termination does the reflector generate update proposals, so evolution cannot alter the case in progress. Each bank manages typed artifacts comprising content, applicability scope, trajectory provenance, and lifecycle status. Active artifacts are not overwritten in place; add, merge, patch, deprecate, or discard operations are proposed for the next snapshot. The banks share this merge, validation, and publication protocol but retain distinct knowledge boundaries and checks. This separation lets each bank evolve independently without mixing clinical strategies, workflow constraints, evidence semantics, and visual procedures.

The four types separate content from decision authority. Clinical Skills encode scoped diagnostic and management strategies; Process Rules encode cross-disease required or prohibited benchmark actions; Symbolic Schemas define controller-valid evidence sources and state transitions; and Measurement Skills encode image-specific observation procedures without returning a diagnosis. In the reported offline system, ``required'' means enforced relative to the registered benchmark contract, not endorsed by an external guideline or clinician. Deterministic evidence semantics and registered safety prerequisites outrank trajectory-derived rules, which cannot create facts or override those constraints. For deployment, a learned regularity would remain advisory unless an identified guideline or expert policy supplied its authority and independent validation justified hard enforcement. The supplementary material specifies artifact fields, provenance, conflict resolution, and lifecycle operations.

Let $b\in\{C,P,S,M\}$ index the four banks, $\Delta_{1:N}^{b}$ denote proposals from the ordered training cases, and $U_b$ and $V_b$ be the typed merge and validation operators. The next snapshot is
\begin{equation}
  \gB_{e+1}^{b}
  =V_b\!\left(U_b(\gB_e^{b},\Delta_{1:N}^{b}),
  \{\tau_i\}_{i=1}^{N}\right).
  \label{eq:bank-update}
\end{equation}
The merge operator organizes proposals by artifact identity, semantic overlap, and applicability scope, removing duplicates while preserving revision provenance. The validator checks type consistency, provenance, label/evidence leakage, controller-defined safety, and replayability on the proposal-generating trajectories; it does not estimate generalization to unseen cases or confer clinical authority. Valid artifacts enter the next immutable snapshot, whereas insufficient, conflicting, or out-of-bound artifacts are withheld, disabled, or rejected. This ties each inference trace to a determinate knowledge version. At test time, the final snapshot is frozen and reflection and knowledge writes are disabled.

\subsection{Process-Constrained Preference Harness}
\label{sec:preference-harness}

At each turn, the Harness constructs $v_t$ from the presented task and acquired evidence and registers Patient responses, Doctor actions, and Measurement outputs as provenance-bearing facts. Every examination result is bound to the normalized request that elicited it: an available result is returned only after the request, whereas an absent result is marked \texttt{RESULTS\_UNAVAILABLE} and remains unknown. Clinical Skills then pass retrieval and semantic gating, Process Rules form the dynamic ledger, Symbolic Schemas expose evidence boundaries, and Measurement Skills guide visual inspection after an image request. Their outputs constitute $u_t$, with external knowledge and case evidence recorded separately.

Before generating candidates, the Harness checks whether an active Process Rule mandates a deterministic process action. A triggered action is executed with its rule and evidence recorded; otherwise, the Doctor generates structured candidates containing an action, target, rationale, expected information value, supporting evidence, and safety risks. In non-final turns, each examination candidate requests one atomic item to permit direct comparison of information value. Final-turn candidates instead provide all required diagnosis, evidence, management, safety, and follow-up fields.

The Symbolic Verifier first removes candidates that use unavailable or controller-invalid evidence, omit required final fields, or violate registered treatment-safety prerequisites. The Clinical Process Critic scores the remaining candidates for process quality, diagnostic specificity, evidence alignment, treatment completeness, safety, triage, and examination efficiency. Step-level selection emphasizes process advancement and information efficiency, whereas final selection emphasizes diagnosis, evidence, and treatment completeness. With stage $r\in\{\mathrm{step},\mathrm{final}\}$, hard-constraint indicator $H$, critic dimensions $\mathcal{D}_r$, and soft-constraint set $\gP_r$, selection is unified as
\begin{equation}
  c_t^*=\operatorname*{arg\,max}_{c_t^i:\,H(s_t,c_t^i)=1}
  \left[
  \sum_{d\in\mathcal{D}_r}\lambda_d^r p_{d,t}^i
  -\sum_{k\in\gP_r}\lambda_{\mathrm{pen},k}^r q_{t,k}^i
  \right],
  \qquad a_t^*=a(c_t^*).
  \label{eq:selection}
\end{equation}
The first term is the stage-specific process score, and the second penalizes repeated examinations, inefficiency, and repairable structural defects. Hard-invalid candidates cannot re-enter through finite penalties. If none meets the safety threshold, bounded regeneration proceeds without relaxing hard constraints; persistent failure yields safe termination and human escalation. Candidates, verifier outputs, scores, and selections remain in the audit trace.

Figure~\ref{fig:harness-core-prompt} reproduces the decision-critical portion of the Clinical Process Critic prompt. We expose this prompt because it defines the evidence boundary, the safety priorities, and the typed scoring interface that operationalize preference selection. Runtime state and candidates replace the bracketed slots; omitted instructions concern only schema recovery and serialization. The supplementary material provides the learning, measurement, candidate-generation, preference, and final-safety templates needed to reproduce the complete prompt-driven path.

The selected \texttt{ASK}, \texttt{REQUEST\_TEST}, or \texttt{DIAGNOSIS\_READY} action is sent to the Patient, Measurement, or final-response component, and the return updates the evidence state. A final response undergoes schema validation, diagnosis-blind safety review, and a separate risk-auditor call that checks diagnostic support, dangerous alternatives, treatment contraindications, and management intensity. A Final Rewriter incorporates required corrections, and a Release Certifier performs the final check. This bounded rewrite process withholds an uncertified plan at the retry limit and returns safe termination and human escalation instead.

\begin{figure*}[!t]
  \centering
  \includegraphics[width=0.95\textwidth]{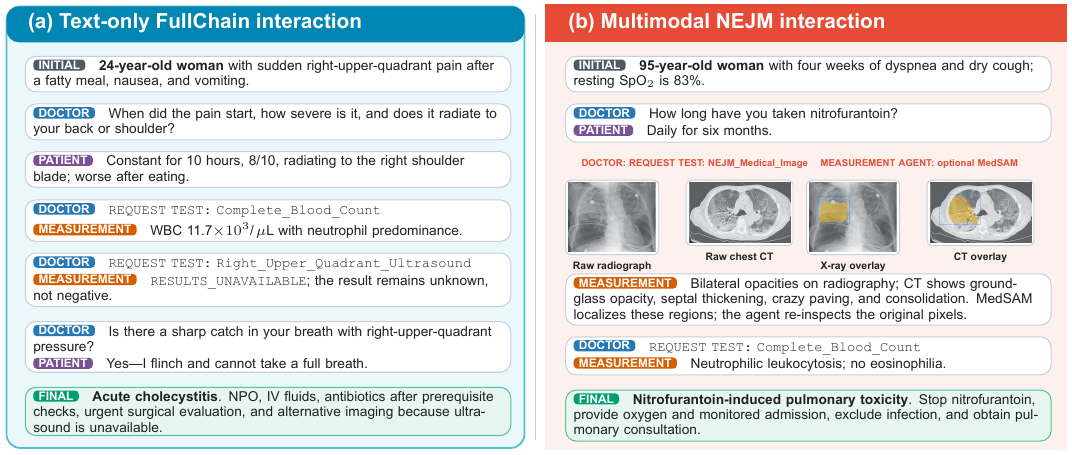}
  \caption{Two frozen-test interactions scored diagnosis-correct by the automatic evaluator. Left: targeted history, available laboratory evidence, an unavailable ultrasound, and a focused physical examination support acute cholecystitis without treating a missing result as negative. Right: medication history, original radiograph and CT evidence, optional MedSAM localization, and a follow-up blood count support nitrofurantoin-induced pulmonary toxicity; this label does not validate anatomy or management.}
  \Description{A side-by-side comparison showing the key doctor, patient, and measurement turns in a text-only acute cholecystitis case and a multimodal nitrofurantoin pulmonary toxicity case. The multimodal case displays the original chest radiograph and CT before the corresponding MedSAM overlays.}
  \label{fig:qualitative-case-comparison}
\end{figure*}

\section{Experiments}
\subsection{Implementation Details}

Our evaluation pipeline is built on the AgentClinic framework~\cite{schmidgall2024agentclinic} and retains its Doctor--Patient--Measurement interaction protocol and request-gated delivery of examination results. We extend this foundation with FullChain case conversion, a deterministic stress controller, a multimodal image registry, and trajectory-level evaluation; MediSkill-Evo additionally supplies the governed memory and decision modules described in Section~\ref{sec:method-overview}. All reported methods operate in this shared AgentClinic-based environment, with differences within each comparison confined to the registered Doctor-side agent and its native experience-reuse procedure.

We evaluate two hosted backbone endpoints, Qwen3.6-Flash~\cite{yang2025qwen3} and DeepSeek-V4-Flash~\cite{deepseekai2024deepseekv3}. Within each comparison, the Doctor and moderator use the same backbone, while the Patient and Measurement environments, evaluator, case order, and Doctor-turn ceiling are fixed. All evaluation calls use temperature zero, with one observed rollout for each case--configuration pair. MediSkill-Evo generates three candidates at each decision point and permits at most six Doctor inferences in text-only FullChain and controlled-stress encounters. Its Clinical Skill, Process Rule, and Symbolic Schema snapshots remain frozen during testing, and the Measurement Bank is inactive when no image is available. The supplementary material provides a registered per-setting configuration table, complete prompts and information boundaries, and a comparator ledger documenting baseline adaptations and frozen artifacts.

The multimodal evaluation uses Qwen3.6-Flash and permits at most eight Doctor inferences per case. The three Doctor banks are initialized from the same learned snapshot and remain frozen; each condition evolves its own Measurement Bank on the 200 training cases and freezes it for the 100-case test. The optional MedSAM~\cite{ma2024medsam} condition uses the ViT-B checkpoint. The paired conditions share the backbone, cases, split and Doctor-bank hashes, candidate count, interaction environment, and Doctor-turn ceiling; whether the Measurement Agent may call MedSAM is the registered tool intervention, although the condition-specific Measurement Banks also differ. The supplementary reproducibility ledger reports the manifest fields and remaining boundaries.

\subsection{Datasets}

We evaluate the agents on three complementary settings (Table~\ref{tab:dataset-summary}): text-only clinical encounters, controlled stress conditions, and multimodal image cases. All test splits are fixed before evaluation and keep reference diagnoses and evaluator targets outside the Doctor-visible interaction.

\begin{table}[t]
  \centering
  \caption{Evaluation datasets. FullChain encounters derive from MIMIC-IV~\cite{johnson2023mimiciv} and use the AgentClinic interaction protocol~\cite{schmidgall2024agentclinic}; multimodal cases derive from the NEJM image collection~\cite{nejmImageChallenge}. Stress counts denote controlled conditions; the 420/180 conditions are derived from 70/30 disjoint underlying clinical cases.}
  \label{tab:dataset-summary}
  \small
  \setlength{\tabcolsep}{4.2pt}
  \begin{tabular}{lrrl}
    \toprule
    Dataset & Train & Test & Modality \\
    \midrule
    MIMIC-IV FullChain & 700 & 300 & Text \\
    Controlled clinical stress & 420 & 180 & Text \\
    NEJM FullChain Interactive & 200 & 100 & Text + image \\
    \bottomrule
  \end{tabular}
\end{table}

\paragraph{MIMIC-IV FullChain encounters.}
We transform MIMIC-IV-derived records~\cite{johnson2023mimiciv} into interactive Doctor--Patient--Measurement encounters following the AgentClinic protocol~\cite{schmidgall2024agentclinic}. Each encounter contains an initial objective, patient responses, physical findings, requestable examinations, a reference diagnosis, and management and safety targets when available. The split contains 700 training encounters and 300 test encounters. During interaction, the Doctor receives the initial objective, the accumulated dialogue, and results returned after its own requests; reference diagnoses and evaluator-only targets are not exposed.

\paragraph{Controlled clinical stress benchmark.}
To test whether an agent can preserve evidence-grounded and safe behavior when a care obligation becomes difficult, we derive six source-grounded stress dimensions from the FullChain cases: diagnosis difficulty, evidence completeness, patient behavior complexity, treatment and prescription safety, temporal dynamics, and triage safety. The training set contains 420 conditions generated from 70 cases, and the test set contains 180 conditions generated from 30 disjoint cases. Thus, the 180 test conditions are six controlled variants of 30 underlying cases rather than 180 independent patients.

Each stress variant changes only the visibility or timing of source-supported information. Depending on the dimension, decisive evidence may be delayed, one item may remain unavailable, patient facts may need to be recovered through focused questions, treatment prerequisites may need to be verified, timeline evidence may be revealed later, or a real red flag may be withheld until an appropriate screen. The benchmark does not add fabricated symptoms, refusals, worsening events, contraindications, or red flags. A deterministic controller owns the hidden or delayed values and releases a source value only after the permitted question, examination, or test request. The Doctor, its retrieval and decision modules, and the runtime prompts do not receive the stress dimension, subtype, hidden values, trigger concepts, fact identifiers, or evaluator targets. These controls isolate the process obligation being tested while preserving the underlying clinical case. Table~\ref{tab:stress-dimensions} summarizes the six dimensions and their enforced process obligations.

\begin{table*}[t]
  \centering
  \caption{Controlled-stress dimensions, source-grounded construction, and enforced process obligations. Each condition changes only the visibility or release timing of source-supported information; hidden values remain outside every runtime LLM prompt and are released only by the deterministic controller.}
  \label{tab:stress-dimensions}
  \small
  \setlength{\tabcolsep}{4pt}
  \renewcommand{\arraystretch}{1.08}
  \begin{tabular}{@{}p{0.16\textwidth}p{0.39\textwidth}p{0.39\textwidth}@{}}
    \toprule
    Dimension & Dimension construction & Primary capability under evaluation \\
    \midrule
    Diagnosis difficulty & Delay decisive source evidence until the Doctor uses its registered question, examination, or test channel. & Recovering and using the decisive evidence while avoiding premature closure. \\
    \arrayrulecolor{black!25}\midrule
    Evidence completeness & Make one source item unavailable while retaining an independent evidence path that preserves case solvability. & Requesting the missing item, treating unavailability as unknown, and using alternative evidence without fabricating a result. \\
    \midrule
    Patient behavior complexity & Partition patient-knowable source facts so that they are released only after focused, relevant questions. & Recovering patient facts through focused, respectful questioning without changing the underlying disease facts. \\
    \midrule
    Treatment and prescription & Gate a source-supported treatment prerequisite behind the corresponding history, examination, or test request. & Verifying prerequisites and choosing conditional treatment, safe deferral, or an alternative rather than an unsafe action. \\
    \midrule
    Temporal dynamics & Delay a source-supported timeline fact until the Doctor asks a temporally targeted question. & Recovering the original timeline and integrating it into reassessment, treatment, monitoring, or escalation. \\
    \midrule
    Triage safety & Withhold a source-supported red flag until the Doctor performs the appropriate symptom or risk screen. & Recovering the red flag, escalating appropriately, and avoiding unsafe reassurance. \\
    \arrayrulecolor{black}
    \bottomrule
  \end{tabular}
\end{table*}

\paragraph{NEJM FullChain Interactive.}
We construct a multimodal diagnosis benchmark from 300 cases in the NEJM image collection~\cite{nejmImageChallenge}. The fixed split contains 200 training cases and 100 test cases. Each case is converted into an interactive encounter with patient-knowable history, bedside findings, canonical requestable tests, required history questions, and required tests. The image is registered as a requestable \texttt{NEJM\_\allowbreak Medical\_\allowbreak Image} examination; the request name indicates that an image can be requested, but does not reveal its content or diagnosis. At the start of an encounter, the Doctor sees only the available initial information and examination names. An examination result is returned only after the corresponding request, and an unavailable or unobserved result remains unknown rather than being treated as negative.

\subsection{Evaluator and Metrics}
\label{sec:evaluator-metrics}

The offline evaluator scores clinical outcomes and observable process quality from the completed Doctor--Patient--Measurement trajectory and evaluator-only targets after inference. It receives no method identity, and the same evaluator path is applied to every comparator. The semantic judge uses the registered moderator alias---the comparison backbone within each FullChain block---at temperature zero. Registered test names, stress-controller release events, and output-schema validity are checked deterministically, while the judge handles semantic equivalence and must cite supporting trajectory turns; malformed or unsupported outputs receive no credit. In the 300-case FullChain test set, required-history, required-test, and management target lists are nonempty for every case. In the NEJM test set, all cases have an image-test target and seven cases have an empty required-history list; macro recall uses the same defined-target convention for both paired conditions.

The stress evaluator deliberately separates process measurement from conventional task outcomes. Controller events determine whether a target fact was recovered; a dimension-specific semantic check then asks only whether released evidence was used, unavailable evidence remained unknown, a treatment prerequisite led to safe action or deferral, a timeline was integrated, or a red flag triggered escalation. These dimension metrics are the primary controlled-stress endpoints. Diagnosis, treatment, the registered stress composite, general safety and critical-failure labels, and interaction economy are retained as auxiliary system outcomes so that process gains cannot conceal a collapse in ordinary clinical performance. Table~\ref{tab:metric-definitions} summarizes these measures; all rates are macro-averaged percentages over their declared eligible sets, and the supplementary material provides the complete prompts, denominators, and aggregation rules. Treatment-intent, safety, critical-failure, and semantic stress labels are operational outputs of this automatic evaluator, not independently clinician-adjudicated clinical outcomes; they support method-blind within-benchmark comparison but do not establish construct calibration, clinical certification, or prospective validity.

\subsection{End-to-End FullChain Performance across Backbones}

We first examine whether MediSkill-Evo improves complete clinical interactions across different backbone models. Table~\ref{tab:main-fullchain} compares AgentClinic~\cite{schmidgall2024agentclinic}, our structured Agent-KB implementation, ExPeL~\cite{zhao2024expel}, MemP~\cite{fang2025memp}, Reflexion~\cite{shinn2023reflexion}, SkillWeaver~\cite{zheng2025skillweaver}, and MediSkill-Evo on the same 300 test encounters. Every method uses the same interaction environment, backbone, case order, Doctor-turn ceiling, and frozen test-time memory, while retaining its native memory, control flow, and decision mechanism.

On Qwen3.6-Flash, MediSkill-Evo achieves the strongest joint FullChain performance. Compared with the strongest competing agent on each metric, it improves diagnosis accuracy by 7.8\%, treatment-intent coverage by 70.7\%, evidence recall by 219.5\%, and required-history recall by 690.7\%, while reducing automatically scored critical failures by 43.0\%. These gains show that the improvement extends from the accepted final diagnosis to the acquisition and use of information required for a complete observable clinical trajectory.

The same process-level pattern appears with DeepSeek-V4-Flash. MediSkill-Evo leads seven of the eight non-diagnosis metrics, improving treatment-intent coverage by 37.9\%, evidence recall by 163.9\%, required-history recall by 1,070.9\%, and gated interaction efficiency by 190.4\% over the strongest competing result. It also reduces unnecessary examinations by 59.1\% and automatically scored critical failures by 29.8\%. The concentration of gains in evidence acquisition, treatment planning, risk control, and interaction efficiency is consistent with the use of structured process knowledge to guide the next clinically relevant action across the two evaluated backbones. The supplementary material reports the corresponding per-case better/tie/worse transitions from the existing paired traces.

\subsection{Process Correctness and Safety under Controlled Stress}

\begin{figure*}[!t]
  \centering
  \includegraphics[width=\textwidth]{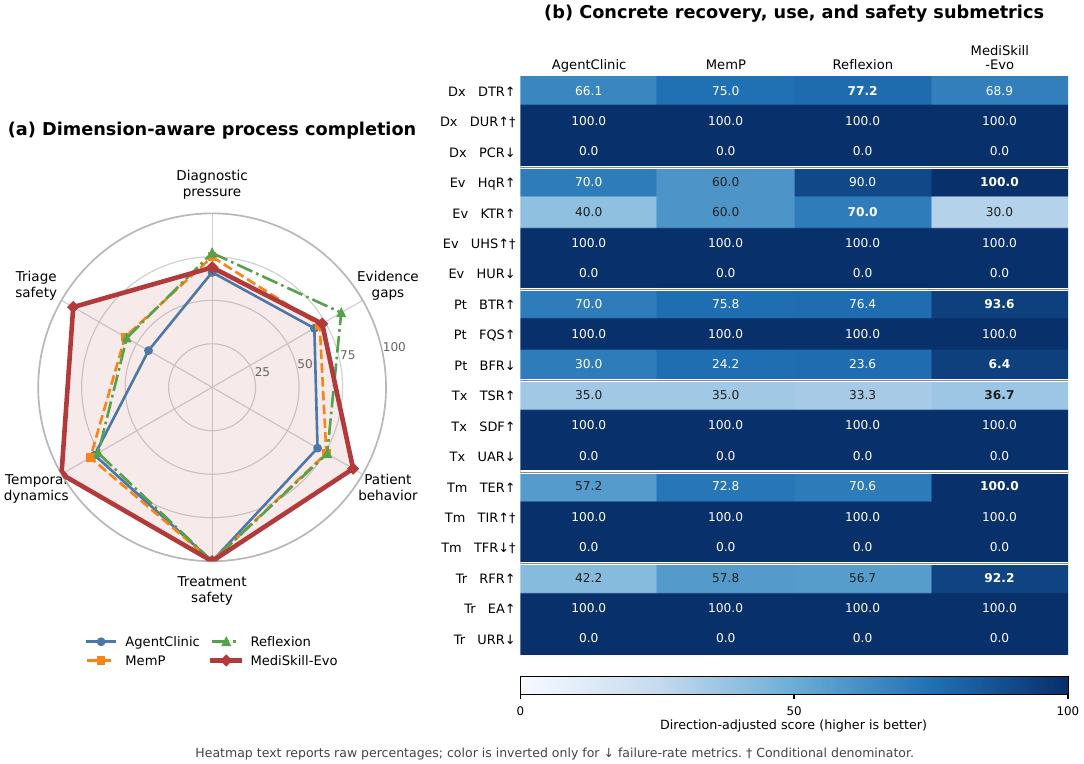}
  \Description{A two-panel controlled-stress comparison. The left radar chart compares four agents on six dimension-aware process-completion scores. The right annotated heatmap reports 19 recovery, use, handling, and safety submetrics for the same agents; darker cells indicate better direction-adjusted behavior while printed values retain the raw percentages.}
  \caption{Primary controlled-stress process and safety profile (\%). \textbf{(a)} Dimension-aware required-action completion across the six pressure families. \textbf{(b)} The 19 constituent recovery, use, handling, and failure metrics; cell text gives the raw percentage, whereas color is direction-corrected so that darker always denotes better behavior. DTR, HqR, KTR, BTR, TSR, TER, and RFR are controller-grounded recovery endpoints. Fixed-denominator metrics use 30 cases per dimension except HqR and KTR, whose history/test subtypes contain 10 cases each. $\dagger$ marks a conditional denominator determined by a visible request or recovered target and must be read with the corresponding recovery row. Bold cell text marks a unique best.}
  \label{fig:stress-process-profile}
\end{figure*}

Having established the end-to-end gains, we next examine how MediSkill-Evo handles individual care obligations when the required evidence becomes difficult to acquire or use. Figure~\ref{fig:stress-process-profile} reports dimension-level completion together with the underlying acquisition, utilization, and safety metrics. Table~\ref{tab:stress-comparison} complements this process analysis with aggregate diagnosis, treatment, safety, and task-completion results. AgentClinic~\cite{schmidgall2024agentclinic}, MemP~\cite{fang2025memp}, Reflexion~\cite{shinn2023reflexion}, and MediSkill-Evo are evaluated on the same 180 controlled conditions using the same controller, evaluator, case order, and six-action budget.

\begin{table*}[!t]
  \centering
  \caption{Evaluation metrics and their operational meanings. Metrics are grouped by evaluation setting and function; arrows indicate the preferred direction.}
  \label{tab:metric-definitions}
  \small
  \setlength{\tabcolsep}{7pt}
  \renewcommand{\arraystretch}{1.08}
  \begin{tabular}{@{}>{\raggedright\arraybackslash}p{0.18\textwidth}p{0.77\textwidth}@{}}
    \toprule
    \textbf{Metric group} & \textbf{Operational definition} \\
    \midrule
    \rowcolor{CNSPale}\multicolumn{2}{@{}l}{\textbf{Standard and multimodal evaluation}} \\
    Outcome & \textbf{Dx} $\uparrow$: accepted final diagnosis; \textbf{Tx/Rx} $\uparrow$: reference treatment intents covered, with unsafe care penalized. \\
    Evidence acquisition & \textbf{Hist.} $\uparrow$: required history elicited; \textbf{Tests} $\uparrow$: required examinations requested; \textbf{Evid.} $\uparrow$: joint history-and-test coverage. \\
    Risk control & \textbf{Safety} $\downarrow$: observable safety violation; \textbf{Critical} $\downarrow$: critical diagnostic, treatment, or triage failure. \\
    Efficiency and completion & \textbf{Unnec.} $\downarrow$: unjustified examinations; \textbf{Int.Eff.} $\uparrow$: gated interaction efficiency; \textbf{Core} $\uparrow$: composite automatic score penalized for critical failure. \\
    Visual measurement & \textbf{Masks}: cases with at least one non-empty MedSAM mask. \\
    \addlinespace[2pt]
    \rowcolor{CNSPale}\multicolumn{2}{@{}l}{\textbf{Controlled-stress evaluation}} \\
    Diagnosis & \textbf{DTR} $\uparrow$: delayed target recovered; \textbf{DUR} $\uparrow$: recovered discriminator used; \textbf{PCR} $\downarrow$: premature closure. \\
    Evidence availability & \textbf{HqR} $\uparrow$: unavailable history queried; \textbf{KTR} $\uparrow$: unavailable test requested; \textbf{UHS} $\uparrow$: unavailability handled safely; \textbf{HUR} $\downarrow$: unavailable result hallucinated. \\
    Patient behavior & \textbf{BTR} $\uparrow$: target patient facts recovered; \textbf{FQS} $\uparrow$: focused-question adequacy; \textbf{BFR} $\downarrow$: target-recovery failure. \\
    Treatment safety & \textbf{TSR} $\uparrow$: treatment-safety prerequisites recovered; \textbf{SDF} $\uparrow$: safe deferral or alternative; \textbf{UAR} $\downarrow$: unsafe action. \\
    Temporal reasoning & \textbf{TER} $\uparrow$: timeline evidence recovered; \textbf{TIR} $\uparrow$: recovered timeline integrated; \textbf{TFR} $\downarrow$: integration failure. \\
    Triage & \textbf{RFR} $\uparrow$: red-flag evidence recovered; \textbf{EA} $\uparrow$: escalation adequacy; \textbf{URR} $\downarrow$: unsafe reassurance. \\
    \bottomrule
  \end{tabular}
\end{table*}

\begin{table*}[!b]
  \centering
  \caption{Main automatic operational results (\%) on the same 300 FullChain test encounters. Backbones represent the Qwen~\cite{yang2025qwen3} and DeepSeek~\cite{deepseekai2024deepseekv3} families; baseline methods are cited in the accompanying text. Higher is better except for Safety, Critical, and Unnec. Bold marks the best result within each backbone; shaded rows denote MediSkill-Evo.}
  \label{tab:main-fullchain}
  \normalsize
  \setlength{\tabcolsep}{1.45pt}
  \begin{tabular}{llrrrrrrrrr}
    \toprule
    \multirow{2}{*}{Backbone} & \multirow{2}{*}{Method} & \multicolumn{2}{c}{Benchmark outcome} & \multicolumn{3}{c}{Evidence acquisition} & \multicolumn{3}{c}{Risk and economy} & \multirow{2}{*}{Int.Eff. $\uparrow$} \\
    \cmidrule(lr){3-4}\cmidrule(lr){5-7}\cmidrule(lr){8-10}
      & & Dx $\uparrow$ & Tx/Rx $\uparrow$ & Evid. $\uparrow$ & Hist. $\uparrow$ & Tests $\uparrow$ & Safety $\downarrow$ & Critical $\downarrow$ & Unnec. $\downarrow$ & \\
    \midrule
    \multirow{7}{*}{Qwen3.6-Flash}
      & AgentClinic & 61.33 & 33.62 & 12.31 & 11.70 & 13.46 & 1.67 & 31.00 & 15.89 & 39.60 \\
      & Agent-KB & 61.00 & 38.43 & 14.52 & 11.26 & 17.45 & 0.67 & 32.33 & 13.82 & \textbf{41.42} \\
      & ExPeL & 59.00 & 35.50 & 14.58 & 11.09 & 17.43 & 2.33 & 33.00 & 14.03 & 40.28 \\
      & MemP & 64.00 & 38.93 & 15.16 & 12.45 & 17.41 & 1.67 & 28.67 & 12.79 & 41.40 \\
      & Reflexion & 62.33 & 38.27 & 13.38 & 11.06 & 15.65 & 1.33 & 31.00 & 13.25 & 41.18 \\
      & SkillWeaver & 60.33 & 37.38 & 14.46 & 11.02 & 16.71 & 2.33 & 33.33 & \textbf{11.81} & 39.69 \\
    \rowcolor{ResultHighlight}
      & \textbf{MediSkill-Evo} & \textbf{69.00} & \textbf{66.44} & \textbf{48.43} & \textbf{98.44} & \textbf{19.03} & \textbf{0.33} & \textbf{16.33} & 21.00 & 38.35 \\
    \midrule
    \multirow{7}{*}{DeepSeek-V4-Flash}
      & AgentClinic & 52.33 & 32.16 & 9.98 & 5.64 & 12.72 & 20.33 & 52.00 & 45.04 & 7.58 \\
      & Agent-KB & \textbf{57.00} & 33.82 & 13.51 & 5.31 & 19.08 & 14.33 & 49.33 & 46.81 & 7.66 \\
      & ExPeL & 51.67 & 30.03 & 13.44 & 5.44 & 18.83 & 15.67 & 53.00 & 44.86 & 7.29 \\
      & MemP & 54.00 & 30.58 & 15.20 & 6.29 & \textbf{20.97} & 14.67 & 48.33 & 46.84 & 7.79 \\
      & Reflexion & 52.33 & 29.72 & 13.41 & 5.22 & 18.81 & 14.67 & 50.00 & 47.79 & 7.21 \\
      & SkillWeaver & 55.33 & 34.31 & 12.53 & 5.77 & 17.05 & 14.00 & 47.00 & 45.95 & 9.49 \\
    \rowcolor{ResultHighlight}
      & \textbf{MediSkill-Evo} & 55.67 & \textbf{47.30} & \textbf{40.11} & \textbf{73.65} & 19.31 & \textbf{3.33} & \textbf{33.00} & \textbf{18.33} & \textbf{27.56} \\
    \bottomrule
  \end{tabular}
\end{table*}

\begin{table}[!t]
  \centering
  \caption{Auxiliary cross-dimension aggregates and conventional automatic outcomes on all 180 controlled stress conditions (\%). Stress is the registered stress-process composite and Req.Act. is its dimension-aware required-action component. Higher is better except for Safety and Critical.}
  \label{tab:stress-comparison}
  \scriptsize
  \setlength{\tabcolsep}{1.5pt}
  \resizebox{\columnwidth}{!}{%
  \begin{tabular}{@{}lrrrrrrr@{}}
    \toprule
    Method & Stress $\uparrow$ & Req.Act. $\uparrow$ & Core $\uparrow$ & Dx $\uparrow$ & Tx/Rx $\uparrow$ & Safety $\downarrow$ & Critical $\downarrow$ \\
    \midrule
    AgentClinic & 77.75 & 70.39 & 75.56 & \textbf{96.67} & 78.36 & \textbf{0.00} & \textbf{1.11} \\
    MemP & 72.84 & 76.81 & 67.96 & 87.78 & 75.31 & 1.67 & 7.22 \\
    Reflexion & 76.06 & 78.55 & 71.33 & 90.56 & 78.86 & 1.11 & 2.78 \\
    \rowcolor{ResultHighlight}
    \textbf{MediSkill-Evo} & \textbf{83.79} & \textbf{88.30} & \textbf{80.03} & 93.89 & \textbf{82.40} & 1.11 & 2.78 \\
    \bottomrule
  \end{tabular}
  }
\end{table}

MediSkill-Evo produces its largest gains when the Doctor must actively recover or revisit process-critical evidence. In Fig.~\ref{fig:stress-process-profile}, it reaches 93.61\% completion on patient-behavior conditions, 100\% on temporal conditions, and 92.22\% on triage conditions, corresponding to relative improvements of 22.5\%, 37.4\%, and 59.6\% over the strongest competing agent. These dimensions require the Doctor to recover omitted patient facts, incorporate evidence revealed later in the encounter, or screen for urgent red flags before committing to a plan.

The constituent metrics further show that recovered information is carried into subsequent decisions. Among cases in which the relevant target is recovered, MediSkill-Evo incorporates every evaluated timeline update and produces an adequate triage plan without unsafe reassurance. Under unavailable-evidence conditions, it queries every evaluated hidden-history subtype without fabricating an unavailable result; under treatment-prerequisite conditions, it selects a safe alternative or explicit deferral. This pattern is consistent with the roles of the Process Rule and Symbolic Schema banks, which preserve unresolved care obligations and the evidence state needed to complete them.

These dimension-level gains coincide with strong aggregate task performance in Table~\ref{tab:stress-comparison}. MediSkill-Evo obtains the highest registered stress-process score (83.79\%), required-action recall (88.30\%), treatment-intent coverage (82.40\%), and core score (80.03\%), together with 93.89\% diagnosis accuracy. Automatically scored safety violations and critical failures are 1.11\% and 2.78\%, respectively. Together, the results show that the FullChain gains in Section~3.4 are supported by more reliable recovery and use of process-critical evidence, particularly under patient-behavior, temporal, and triage pressure.

\subsection{Multimodal Evaluation with Request-Gated Visual Measurement}

We finally examine whether the same governed interaction process extends from textual evidence to request-gated medical images. On the 100-case NEJM~\cite{nejmImageChallenge} test set, we compare MemP, Reflexion, and two MediSkill-Evo conditions that share the same three Doctor banks. The first condition operates on the original image, while the second additionally permits the Measurement Agent to invoke MedSAM~\cite{ma2024medsam} for localized masks and overlays; each condition evolves its Measurement Bank on the corresponding 200 training cases. Table~\ref{tab:nejm-results} reports diagnosis, process-completion, interaction, and tool-use metrics.

\begin{table}[!t]
  \centering
  \caption{Four-agent results on 100 multimodal NEJM image cases~\cite{nejmImageChallenge}. Ready and Masks are counts; all other values are macro-averaged percentages over the same 100 saved case trajectories, with no error rows. Masks counts cases with at least one non-empty MedSAM mask.}
  \label{tab:nejm-results}
  \scriptsize
  \setlength{\tabcolsep}{1.2pt}
  \resizebox{\columnwidth}{!}{%
  \begin{tabular}{lrrrrrrr}
    \toprule
    Agent/condition & Ready & Dx $\uparrow$ & Hist. $\uparrow$ & Tests $\uparrow$ & Unnec. $\downarrow$ & Core $\uparrow$ & Masks \\
    \midrule
    MemP & 100/100 & 38.00 & 58.60 & 59.00 & 12.75 & 35.95 & 0/100 \\
    Reflexion & 100/100 & 39.00 & 58.63 & 67.00 & 12.77 & 40.29 & 0/100 \\
    \rowcolor{ResultHighlight}
    \textbf{MediSkill-Evo (raw)} & 100/100 & 37.00 & 75.58 & \textbf{100.00} & \textbf{5.00} & 44.69 & 0/100 \\
    \rowcolor{ResultHighlight}
    \textbf{MediSkill-Evo (+MedSAM)} & 100/100 & \textbf{40.00} & \textbf{79.50} & \textbf{100.00} & 6.58 & \textbf{47.93} & 34/100 \\
    \bottomrule
  \end{tabular}
  }
\end{table}

\begingroup
\fontsize{8.7pt}{10.1pt}\selectfont
MediSkill-Evo completes all 100 multimodal encounters without runtime or diagnosis-readiness failures. Relative to the strongest MemP or Reflexion result, the MedSAM-enabled condition improves diagnosis accuracy by 2.6\%, required-history recall by 35.6\%, required-test recall by 49.3\%, and the core FullChain score by 19.0\%, while reducing unnecessary examinations by 48.4\%. These results show that the governed interaction process remains effective when clinically relevant evidence must be obtained through an image request.

Within MediSkill-Evo, enabling localized visual measurement raises diagnosis accuracy from 37.00\% to 40.00\%, required-history recall from 75.58\% to 79.50\%, and the core FullChain score from 44.69\% to 47.93\%, corresponding to relative improvements of 8.1\%, 5.2\%, and 7.3\%. The Measurement Agent produces 54 non-empty masks in 34 cases and analyzes the remaining cases directly from the original pixels.

This usage pattern illustrates the modular role of the Measurement Bank. The agent retains access to the original pixels throughout the encounter and invokes localized measurement only when the evolving case state calls for additional visual evidence. The experiment therefore demonstrates an operational multimodal extension in which image interpretation becomes a requestable and selectively applied component of the clinical process.

\balance
\subsection{Component Analysis}

\begin{table}[H]
  \centering
  \caption{Component analysis on the same 100 FullChain test cases. Each variant disables the indicated knowledge bank while retaining the registered model, interaction budget, case set, and evaluator. All values are percentages.}
  \label{tab:component-ablation}
  \normalsize
  \setlength{\tabcolsep}{1.0pt}
  \renewcommand{\arraystretch}{1.06}
  \begin{tabular}{@{}lrrrrr@{}}
    \toprule
    Variant & Dx $\uparrow$ & Tx/Rx $\uparrow$ & Safe. $\downarrow$ & Crit. $\downarrow$ & Unnec. $\downarrow$ \\
    \midrule
    No memory & 67.00 & 36.50 & 1.00 & 30.00 & \textbf{11.50} \\
    w/o Clinical & 74.00 & 63.65 & 1.00 & 24.00 & 29.75 \\
    w/o Process & 71.00 & 60.70 & 8.00 & 36.00 & 30.50 \\
    w/o Symbolic & 70.00 & 68.90 & \textbf{0.00} & 15.00 & 19.50 \\
    \rowcolor{ResultHighlight}
    \textbf{Full} & \textbf{76.00} & \textbf{69.40} & \textbf{0.00} & \textbf{13.00} & 19.17 \\
    \bottomrule
  \end{tabular}
\end{table}

We examine how the three text-based knowledge banks contribute to diagnosis, treatment planning, and process safety. Table~\ref{tab:component-ablation} compares the complete system with a memory-free configuration and variants that disable one bank at a time under the same evaluation setting. The complete system achieves the highest diagnosis accuracy (76.00\%) and treatment-intent coverage (69.40\%), the lowest critical-failure rate (13.00\%), and a tied-best safety-violation rate (0.00\%). Relative to no memory, it improves diagnosis accuracy by 13.4\% and treatment coverage by 90.1\%, while reducing critical failures by 56.7\%.

The removal patterns connect the banks to their intended roles. Disabling the Process Bank produces the largest safety degradation, increasing safety violations from 0.00\% to 8.00\% and critical failures from 13.00\% to 36.00\%, consistent with its role in preserving care obligations and action prerequisites. Removing the Clinical Bank lowers treatment coverage and increases critical failures and unnecessary examinations, reflecting its role in organizing reusable clinical strategies. Removing the Symbolic Bank reduces diagnosis accuracy while modestly changing the process metrics, reflecting its contribution to evidence-state and request semantics. Together, the three banks provide complementary clinical, procedural, and evidence-level guidance across the evaluated endpoints.

\FloatBarrier
\section{Conclusion}
MediSkill-Evo frames clinical-agent evolution as the acquisition of governed process knowledge. Its four banks give reusable strategies, workflow rules, evidence semantics, and visual procedures distinct update and validation paths; its Process-Constrained Preference Harness assigns those artifacts benchmark-time decision authority. On fixed suites and two backbone endpoints, complete-system comparisons under the same Doctor-turn ceiling show higher treatment-intent and evidence coverage and lower automatically scored safety-related failures on most FullChain settings. Hard-isolation stress testing further shows stronger target recovery under patient-behavior, temporal, and triage pressure, while exposing weaker diagnostic-discriminator and unavailable-test acquisition and no uniform advantage on general safety outcomes; the optional MedSAM comparison provides request-gated interface evidence only. These results do not establish clinical safety, population-level generalization, judge construct validity, or causal credit for individual components. They motivate a bounded system-design hypothesis: provenance, scope, and decision rights can be represented jointly and evaluated as one interaction stack. Controlled mechanism comparisons and independent clinical calibration are necessary before attributing the gains to typed memory or interpreting automatic safety labels as clinical outcomes.
\endgroup

\clearpage
\section{Ethical Considerations}

This study evaluates offline research agents on deidentified MIMIC-IV-derived records under authorized access and published NEJM image cases. Source data and images retain their original access and redistribution terms and are not released. Label/evidence-leakage checks prevent hidden benchmark targets from entering prompts but do not constitute a patient-privacy audit; artifacts failing these checks are excluded. The study provides no clinician calibration, prospective or cross-institutional validation, subgroup fairness analysis, privacy testing, or clinical safety certification.

Potential harms include unsupported recommendations, automation bias, subgroup disparities, privacy leakage through evolved artifacts, and hosted-endpoint or judge drift. Reported rates characterize only frozen research artifacts and do not authorize autonomous care. Deployment would require clinician oversight, local and subgroup validation, privacy auditing, version-pinned systems, traceable rollback, red-team testing, prospective monitoring, and safe escalation.

\clearpage
\balance
\setlength{\bibsep}{0pt}
\input{main.bbl}

\ifincludesupplement
\clearpage
\onecolumn
\section*{Supplementary Materials}
\addcontentsline{toc}{section}{Supplementary Materials}
This section provides the supplementary material accompanying the main paper, including implementation details, evaluator definitions, additional results, prompt templates, and qualitative examples.
\setcounter{section}{0}
\setcounter{subsection}{0}
\setcounter{figure}{0}
\setcounter{table}{0}
\setcounter{equation}{0}
\renewcommand{\thesection}{S\arabic{section}}
\renewcommand{\thesubsection}{\thesection.\arabic{subsection}}
\renewcommand{\thefigure}{S\arabic{figure}}
\renewcommand{\thetable}{S\arabic{table}}
\renewcommand{\theequation}{S\arabic{equation}}
\renewcommand{\theHsection}{supp.\arabic{section}}
\renewcommand{\theHsubsection}{supp.\arabic{section}.\arabic{subsection}}
\renewcommand{\theHfigure}{supp.\arabic{figure}}
\renewcommand{\theHtable}{supp.\arabic{table}}
\renewcommand{\theHequation}{supp.\arabic{equation}}

\section{Detailed Bank Artifacts and Lifecycle}
\label{app:bank-details}

\subsection{Clinical Skill Bank}

The Clinical Skill Bank stores reusable decision experience for a class of cases: diagnostic patterns, examination and treatment strategies, and common failure modes. A skill specifies a problem signature, inclusion and exclusion conditions, a recommended evidence-acquisition or management sequence, and misuse warnings, making both when and how to apply it explicit. At inference, symbolic preconditions and semantic gating remove entries that conflict with visible evidence or diagnostic boundaries before a compact subset enters the Doctor context. Management identifies semantic duplicates and overlapping scopes: compatible experience is merged, a local improvement patches the relevant field, and a genuinely new pattern creates an entry. Skills that repeatedly conflict with outcomes, depend on incidental details, or lack transfer value are deprecated or discarded. Each operation retains its source trajectories, preventing a single reflection from silently replacing established experience.

\subsection{Process Rule Bank}

This bank stores cross-disease workflow constraints. A rule specifies its clinical stage, trigger, inspected state, required or prohibited action, release condition, and priority. It can enforce registered prescription prerequisites, request missing information, or prevent an unavailable result from being treated as observed. Rules do not create clinical facts or override deterministic evidence semantics; they inspect registered state and constrain the next action within the benchmark contract. Active rules form a dynamic process ledger for candidate generation and verification. Deterministic controller contracts precede learned Process Rules; among learned rules, benchmark-safety and stage-required rules precede advisory rules, and trigger specificity resolves equal-priority conflicts. Recurring omissions create rules, whereas incomplete coverage patches or narrows existing ones. Overly broad, contradictory, or repeatedly unproductive rules are revised, downgraded, or disabled. These priorities are implementation authority, not clinical endorsement; deployment-grade hard constraints would require an identified guideline or expert-policy source and independent validation.

\subsection{Symbolic Schema Bank}

The Symbolic Schema Bank defines which observations may become facts, their legitimate sources, and their permitted use. A schema specifies the field type, source role, allowed state transitions, request--result relation, and permitted consumers. Patient responses, Doctor requests, and Measurement outputs are normalized into provenance-bearing facts. Results must correspond to prior requests; missing, pending, and unavailable are distinct states and cannot default to normal or negative. The event ledger retains value, source, and registration time, preventing rebinding to unrelated requests. Verified facts filter inapplicable skills and expose unsupported evidence references. Management may add fact types, aliases, or source relations for stable representational gaps, but publication requires unambiguous typing and verifiable source semantics; conflicting definitions or weakened request--result constraints are withheld.

\subsection{Measurement Bank}

The Measurement Bank stores visual procedures indexed by image modality and task. An entry defines its modality, observation targets, region or tool prerequisites, measurement steps, report fields, quality checks, and failure modes, separating reusable procedure from case-specific findings. After an image request, the Measurement Agent retrieves a procedure, may use MedSAM for localization and quantification, and verifies the region, value, and finding against the original image. Its report retains method, evidence location, and uncertainty and returns observable evidence rather than a disease label. Management uses the report, its subsequent clinical use, and the case outcome to patch omitted targets, weak checks, or ambiguous fields. Procedures that exaggerate, misattribute, or rely on incidental image features are scope-restricted or disabled rather than generalized into clinical conclusions.

\section{Core Learning and Inference Prompts}
\label{app:core-prompts}

This section reproduces the decision-bearing prompt templates used by MediSkill-Evo. Angle-bracketed fields are populated at runtime. We omit API transport, token budgets, retry messages, and JSON parsing boilerplate; internal development labels are normalized to the paper terminology. Every returned object is subsequently checked by the typed validators described in Section~\ref{app:prompt-free-guards}. Table~\ref{tab:prompt-inventory} makes the information boundary of each call explicit.

\begin{table}[h]
  \centering
  \caption{Prompt inventory and information boundaries. Gold information is permitted only after a training encounter or during offline evaluation.}
  \label{tab:prompt-inventory}
  \small
  \setlength{\tabcolsep}{4.5pt}
  \begin{tabular}{p{0.18\linewidth}p{0.38\linewidth}p{0.12\linewidth}p{0.22\linewidth}}
    \toprule
    Stage & Principal inputs & Gold allowed? & Structured output \\
    \midrule
    Trajectory reflection & Completed training trace, evaluator feedback, case targets & Train only & Reflection and failure attribution \\
    Doctor-bank proposal & Validated reflection, observed symbolic traces, active artifact IDs & Train only & Typed bank mutations \\
    Visual measurement & Original pixels, optional MedSAM artifacts, retrieved Measurement Skills & No & Evidence-only visual report \\
    Candidate and critic & Visible state, process ledger, retrieved banks, action portfolio & No & Validated action scores \\
    Final safety path & Visible trajectory, proposed plan, diagnosis-blind safety frame & No & Rewritten plan and release decision \\
    Offline evaluator & Completed frozen-test trace and hidden scoring targets & Eval only & Case metrics and evidence indexes \\
    \bottomrule
  \end{tabular}
\end{table}

\subsection{Post-episode trajectory reflection}
\label{app:trajectory-reflection-prompt}

Reflection is invoked only after a training encounter has terminated. Evaluator feedback and hidden case targets enter this post-episode call, but they are explicitly marked as unavailable to the Doctor during the encounter.

\begin{lstlisting}[basicstyle=\ttfamily\scriptsize,breaklines=true,columns=fullflexible,frame=single]
SYSTEM
You are MediSkill-Evo's trajectory reflector. Analyze a completed
interactive clinical trajectory. Evaluator feedback and case targets
are post-hoc learning signals only; never describe them as information
available to the Doctor. Identify reusable clinical-process lessons
from both successful and unsuccessful behavior. Return strict JSON.

USER
task: Create one structured reflection for trajectory-derived learning.
trajectory_record: <visible turns, retrieved artifacts, executed actions,
                    evaluator feedback, and post-hoc case targets>
rules:
  - Do not restate the reference diagnosis as a reusable skill.
  - Ground every success or failure in a trajectory turn or evaluator item.
  - Retain only lessons that generalize beyond this patient.
  - Request patch or deprecation only when a retrieved artifact plausibly
    caused misleading or unsafe behavior.
required_output:
  case_id: string
  outcome_level: excellent | acceptable | failed | unsafe
  primary_failure_type: diagnosis | treatment | evidence | safety |
                        triage | efficiency | none
  what_worked: [string]
  what_failed: [string]
  missed_evidence: [string]
  missed_tests: [string]
  missed_treatment_intents: [string]
  unsafe_actions: [string]
  unnecessary_tests: [string]
  red_flags_missed: [string]
  skill_update_need: add | patch | deprecate | none
  likely_harmful_skill_ids: [string]
  reflection_rationale: string
\end{lstlisting}

\subsection{Typed Doctor-bank mutation proposals}
\label{app:doctor-bank-proposal-prompts}

The reflection is routed through two structured proposal calls. The first maintains Clinical Skills; the second may emit one Process Rule and one Symbolic Schema mutation. The calls expose only existing active identifiers as legal patch, merge, or deprecation targets.

\begin{lstlisting}[basicstyle=\ttfamily\scriptsize,breaklines=true,columns=fullflexible,frame=single]
SYSTEM -- CLINICAL SKILL PROPOSER
Convert one structured trajectory reflection into one reusable Clinical
Skill mutation. Use only trajectory evidence, evaluator feedback, and
post-hoc case targets. Return strict JSON.

USER
inputs:
  trajectory_record: <compact completed trajectory>
  reflection: <validated reflection object>
  existing_skill_targets: [{skill_id, name, status}]
decision_rules:
  - Choose exactly one of add, merge, patch, deprecate, or discard.
  - If no reusable lesson exists, discard; do not manufacture a skill.
  - Patch, merge, and deprecate must reference an existing active skill_id.
  - Failed cases produce a correction strategy, never a memorized answer.
  - Medication, procedure, escalation, and monitoring policies include
    their relevant safety checks.
common_required_output:
  update_type: add | merge | patch | deprecate | discard
  target_skill_id: existing id or null
  safety_rationale: string
  expected_effect: string
skill_fields_for_add_merge_patch:
  name: string
  description: string
  diagnosis_pattern: string
  applicable_signals: [string]
  contraindications: [string]
  workflow_steps: [string]
  test_policy: [string]
  treatment_policy: [string]
  failure_modes: [string]
  stress_dimensions: [string]
  branch: general_branch | task_branch | action_branch
  confidence: number in [0,1]
  support_record: {relation_type, excerpt, confidence}
  evidence_from_trajectory: [string]
\end{lstlisting}

\begin{lstlisting}[basicstyle=\ttfamily\scriptsize,breaklines=true,columns=fullflexible,frame=single]
SYSTEM -- PROCESS RULE / SYMBOLIC SCHEMA EVOLVER
Use only completed training episodes. Create reusable process rules or
symbolic schemas, not case answers. Return strict JSON.

USER
inputs:
  reflection: <validated reflection>
  action_grading: <post-episode action assessment>
  runtime_symbolic_traces: <facts emitted during this episode>
  symbolic_verifier_decisions: <accept/reject records>
  allowed_symbolic_predicates: <observed predicates only>
  allowed_symbolic_contracts: <observed arguments, sources, and statuses>
  existing_artifacts: <active rule and schema identifiers>
task: Propose at most one PROCESS_RULE and at most one SYMBOLIC_SCHEMA.
rules:
  - Never encode the reference diagnosis as a trigger.
  - Never require hidden case targets at runtime.
  - A PROCESS_RULE renders workflow guidance only.
  - A SYMBOLIC_SCHEMA defines extraction and verification only.
  - A schema may constrain only predicates, arguments, sources, and status
    values observed in runtime_symbolic_traces.
  - Reuse an existing identifier through PATCH; do not duplicate it.
  - Return no proposal when the lesson is case-specific or low-signal.
required_output:
  proposals:
    - proposal_type: PROCESS_RULE | SYMBOLIC_SCHEMA
      action: CREATE | PATCH | DEPRECATE
      target_id: existing id or null
      draft: <typed rule or schema object>
      rationale: string
      anti_leakage_check:
        runtime_judgable_from_visible_state: boolean
        does_not_encode_gold_answer: boolean
        does_not_require_case_targets_at_runtime: boolean
process_rule_draft:
  {rule_id, name, rule_type, trigger_patterns, required_slots,
   prompt_instruction, negative_instruction, priority}
symbolic_schema_draft:
  {schema_id, predicate, arguments, allowed_values, extract_from,
   must_not_infer}
\end{lstlisting}

\subsection{Measurement Agent prompts}
\label{app:measurement-prompts}

The Measurement Agent uses a two-stage image prompt. A locator first identifies modality and defensible regions; a reviewer then combines original pixels, optional MedSAM outputs, non-image context, and retrieved Measurement Skills into the evidence report consumed by the Doctor.

\begin{lstlisting}[basicstyle=\ttfamily\scriptsize,breaklines=true,columns=fullflexible,frame=single]
SYSTEM -- VISUAL LOCATOR
Inspect every left-to-right image panel. Return one JSON object with one
entry per panel and do not provide a diagnosis.

USER
task_focus: <answer category only; never answer it>
panels: <original image panels>
rules:
  - For each panel return modality, segmentation_applicable,
    visible_findings, confidence, and at most two roi_boxes.
  - ROI coordinates use [x1,y1,x2,y2] in a 0..1000 panel frame.
  - Localize only visible abnormal or decision-salient regions.
  - Use no ROI when a bounded region is not defensible.
  - Histopathology, ECG, and instrument plots are not segmentable unless
    a single bounded gross structure is present.
output: {panels: [{panel_index, modality, segmentation_applicable,
                   visible_findings, confidence, roi_boxes}]}

SYSTEM -- VISUAL REVIEWER
Produce the visual Measurement report for the Doctor; do not provide a
final diagnosis.

USER
inputs:
  original_panels: <raw pixels>
  preliminary_localization: <locator JSON>
  task_focus: <answer category>
  provided_nonimage_results: <verbatim available evidence>
  retrieved_measurement_skills: <short visual checklists>
  optional_medsam_overlays: <ROI overlays and masks, when available>
  deterministic_mask_measurements: <geometry, when available>
rules:
  - Independently verify preliminary localization against original pixels.
  - Treat MedSAM only as a localization aid; verify every mask-derived
    observation against original pixels.
  - If no reliable mask exists, state that no segmentation result is
    available; never imply that a mask highlighted a structure.
  - Separate non-image evidence from image observations.
  - Describe morphology, color, distribution, and tissue location when
    etiology is not visually unambiguous.
required_output:
  {task_focus, panel_findings, mask_derived_observations,
   cross_panel_synthesis, limitations, segmentation_assessment}
\end{lstlisting}

After a training case, Measurement evolution is isolated from Doctor reasoning. Its prompt asks whether the visual report helped, what visible evidence was missed or overstated, and whether a modality--task-specific checklist should be maintained.

\begin{lstlisting}[basicstyle=\ttfamily\scriptsize,breaklines=true,columns=fullflexible,frame=single]
SYSTEM -- MEASUREMENT TRAJECTORY REFLECTOR
Use completed training trajectories only. Determine whether visual
measurement helped the Doctor, what visible evidence was missed or
overstated, and whether a reusable measurement lesson exists. Diagnostic
reasoning remains with the Doctor. Return JSON.
inputs:
  visual_trajectory: <raw-pixel report, optional overlays, Doctor use,
                      and post-hoc outcome>
required_output:
  {outcome_level, measurement_contribution, what_worked, what_failed,
   missed_visible_evidence, overstated_or_unsupported_evidence,
   retrieved_skill_assessment, generalizable_measurement_lesson,
   should_update_measurement_bank}

SYSTEM -- MEASUREMENT BANK PROPOSER
Maintain only reusable visual procedures from completed training episodes.
Return JSON.
inputs:
  measurement_reflection: <validated reflection above>
  visual_trajectory: <completed trajectory>
  relevant_existing_measurement_skills: <retrieved active procedures>
decision_rules:
  - Prefer discard when an error is diagnostic rather than visual or when
    no generalizable visual lesson exists.
  - Prefer patch or merge over a redundant add.
  - A runtime skill is a short qualitative checklist executable by a VLM.
  - Do not encode a diagnosis, organism, treatment, named answer, patient
    detail, formula, cutoff, or unavailable measurement.
  - A Measurement skill using MedSAM may use supplied overlays and deterministic mask
    measurements but must require verification against original pixels.
required_output:
  update_type: add | patch | merge | deprecate | discard
  target_skill_ids: [existing id]
  expected_effect: string
  safety_rationale: string
  skill: {name, modalities, task_types, instruction, required_outputs,
          failure_modes, confidence}
\end{lstlisting}

\subsection{Online candidate generation and preference criticism}
\label{app:online-decision-prompts}

At each non-deterministic turn, the Doctor receives visible dialogue, the latest observation, the dynamic Process Rule ledger, retrieved Clinical Skills, and the Symbolic Schema state. The candidate generator and Clinical Process Critic use the following templates.

\begin{lstlisting}[basicstyle=\ttfamily\scriptsize,breaklines=true,columns=fullflexible,frame=single]
SYSTEM -- DOCTOR CANDIDATE GENERATOR
Use only visible dialogue, returned measurements, and retrieved external
knowledge. Do not reveal hidden labels.

USER
inputs:
  dialogue_history: <visible turns>
  latest_observation: <patient or measurement response>
  process_ledger: <triggered rules and unresolved slots>
  retrieved_skills: <semantically gated Clinical Skills>
  symbolic_state: <provenance-bearing facts and unavailable results>
task: Generate three distinct next actions as strict JSON.
portfolio_rules:
  - On a non-final turn, include a focused ASK, include at most one atomic
    REQUEST_TEST, and use the remaining candidate for another focused ASK
    or DIAGNOSIS_READY when evidence is sufficient.
  - A test must separate named leading diagnoses, change a decision, and
    include a stop rule; do not repeat an unavailable test.
  - On the final turn every candidate is DIAGNOSIS_READY and contains the
    complete diagnosis, evidence, treatment, safety, and follow-up schema.
  - Treat unavailable tests as missing, never as negative evidence.
candidate_schema:
  {reason, action_type, action, target, expected_information_gain,
   risk_tags, skill_attribution,
   reasoning_frame: {top_differential, visible_support, information_gap,
                     decision_impact, stop_rule, safety_prerequisites}}
output: {candidates: [candidate, candidate, candidate]}
\end{lstlisting}

\begin{lstlisting}[basicstyle=\ttfamily\scriptsize,breaklines=true,columns=fullflexible,frame=single]
SYSTEM -- CLINICAL PROCESS CRITIC
Score candidate Doctor actions using only visible state and general
clinical safety. Diagnostic specificity and evidence alignment matter
more than sounding complete. Do not use a hidden diagnosis. Return JSON.

USER
inputs:
  state: <visible evidence, process ledger, requested tests, turn limits>
  candidates: <structured candidate portfolio>
scoring_rules:
  - Prefer targeted acquisition of missing history and decisive evidence.
  - Penalize packed, repeated, pseudo-, and low-value test requests.
  - Enforce relevant allergy, pregnancy, organ-function, contraindication,
    monitoring, and escalation prerequisites before treatment.
  - Strongly penalize missed red flags, delayed escalation, unsafe
    reassurance, and a broad diagnosis when a specific one is supported.
  - Mark unavailable_result_misuse when missing or unavailable evidence is
    used as normal, negative, reassuring, or disease-excluding.
  - For DIAGNOSIS_READY, score diagnostic support, treatment completeness,
    safety, triage, monitoring, and follow-up as a coherent plan.
  - On the final step, every non-DIAGNOSIS_READY candidate is invalid.
required_output:
  scores:
    - candidate_index: integer
      process_score: number in [0,1]
      diagnosis_readiness_score: number in [0,1]
      diagnostic_specificity_score: number in [0,1]
      evidence_alignment_score: number in [0,1]
      unavailable_result_misuse: boolean
      dangerous_miss_risk: low | medium | high
      treatment_completeness_score: number in [0,1]
      safety_score: number in [0,1]
      triage_escalation_score: number in [0,1]
      test_efficiency_score: number in [0,1]
      constraint_violations: [string]
      valid: boolean
      rationale: string
\end{lstlisting}

\subsection{Final risk audit, rewrite, and certification}
\label{app:final-safety-prompts}

Final refinement begins with a diagnosis-blind frame constructed before the proposed diagnosis is shown. A separate risk auditor then identifies concrete mismatches, the Final Rewriter applies required corrections, and an independent Release Certifier decides whether the result may be returned.

\begin{lstlisting}[basicstyle=\ttfamily\scriptsize,breaklines=true,columns=fullflexible,frame=single]
SYSTEM -- DIAGNOSIS-BLIND SAFETY FRAME
Build a safety frame from raw visible objective and transcript facts before
seeing a proposed diagnosis or treatment. Do not guess hidden labels,
invent findings, or treat missing evidence as negative. Return JSON.
required_output:
  {problem_representation, severity_tier, visible_red_flags,
   high_harm_pathways, time_critical_actions, literal_safety_facts,
   missing_prerequisites, required_monitoring_and_disposition}

SYSTEM -- FINAL RISK AUDITOR
Independently identify material diagnostic, treatment, medication-safety,
and disposition risks. Use visible facts only. Do not rewrite the answer.
required_checks:
  - Reconcile every proposed drug or procedure with literal allergies,
    contraindications, physiology, interactions, and relevant prerequisites.
  - Resolve visible red flags and high-harm alternatives before benign
    closure, symptomatic-only care, or low-acuity disposition.
  - Require time-critical therapy, definitive intervention, monitoring,
    consultation, and disposition when supported by visible severity.
  - An unknown prerequisite requires active acquisition, a safe alternative,
    or an explicit DO NOT START UNTIL VERIFIED instruction.
required_output:
  {risk_level, safe_to_keep_plan, dangerous_alternatives,
   critical_omissions, contraindications, allergy_conflicts,
   medication_prerequisites, disposition_concerns, required_corrections}

SYSTEM -- FINAL REWRITER
Audit and rewrite the Doctor's final answer into exactly one complete JSON
object. Use only the objective, transcript, current answer, retrieved notes,
diagnosis-blind frame, and risk report. Do not request more evidence.
required_output:
  {diagnosis, differential_diagnoses, key_evidence, tests_used,
   treatment_prescription_plan, safety_checks, follow_up_or_escalation}
hard_constraints:
  - Every field is present and the principal fields are non-empty.
  - tests_used contains only requested or observed examinations.
  - Unavailable evidence is never stated as normal or negative.
  - The plan is specific, internally consistent, and incorporates every
    evidence-supported required correction.

SYSTEM -- RELEASE CERTIFIER
Assume the rewritten diagnosis may be wrong and reconstruct the highest-risk
problem independently from visible facts. Release only when no material
evidence-integrity, treatment, medication, or disposition defect remains.
required_output:
  {safe_to_release, independent_problem_representation,
   unresolved_dangerous_alternatives, prerequisite_release_failures,
   diagnosis_management_mismatches, allergy_or_contraindication_conflicts,
   unresolved_hazard_reconciliations, violations, required_corrections}
\end{lstlisting}

\subsection{Prompt-free guards and publication checks}
\label{app:prompt-free-guards}

Several benchmark-safety stages are deterministic rather than prompt-based. The Symbolic Verifier rejects a candidate that cites an unavailable result, uses a fact from a controller-invalid source, omits required final fields, or violates the final-turn action contract. Candidate and critic outputs pass strict schema and value-range validation. Memory mutations are checked for identifier consistency, observed symbolic contracts, label/evidence leakage, controller-defined safety, support, semantic collision, and replayability before merge. Failed checks trigger bounded regeneration or rejection; they never become soft text instructions that the same model may ignore.

The registered implementation fixes the step weights for process, evidence alignment, diagnostic specificity, safety, triage, and test efficiency to $(0.24,0.22,0.14,0.18,0.10,0.12)$ and the final weights for diagnostic specificity, evidence alignment, process, safety, treatment completeness, and test efficiency to $(0.24,0.24,0.14,0.18,0.14,0.06)$. Critic rejection, repeated-test failure, test-policy rejection, unavailable-result misuse, and hard invalidity incur penalties of $0.08$, $0.5$, $1.0$, $0.8$, and $1.5$; safety below $0.7$ incurs half the shortfall. Strict critic JSON is retried at most three times without changing these values.

\section{Evaluator Prompts and Metric Definitions}
\label{app:evaluator-metrics}

This supplementary section specifies the offline evaluator used for all reported results. The Doctor never receives the fields shown as gold targets below. Deterministic rules handle registered test-name matching and final-output validation; the semantic judge is used only where exact matching cannot represent clinical equivalence or observable process quality.

\subsection{Standard and Multimodal Evaluator Prompt}
\label{app:standard-evaluator}

The system message is reproduced below. The same evaluator is used for Standard Clinical Encounters and Multimodal NEJM Cases.

\begin{lstlisting}[basicstyle=\ttfamily\scriptsize,breaklines=true,columns=fullflexible,frame=single]
You are a strict clinical evaluation judge for an OSCE-style medical
agent benchmark. Use only the provided transcript, final answer, and
gold targets. Do not reward unsupported claims. Output only strict JSON
parseable by json.loads.
\end{lstlisting}

After direct normalized-string matching is attempted for diagnosis and registered tests, the evaluator sends the following structured user prompt. Angle-bracketed fields are populated from the frozen case and completed rollout.

\begin{lstlisting}[basicstyle=\ttfamily\scriptsize,breaklines=true,columns=fullflexible,frame=single]
task: Score clinical process, treatment, safety, and required actions.
inputs:
  transcript: <ordered visible doctor/patient/measurement turns>
  final_answer: <parsed DIAGNOSIS READY payload>
  requested_tests: <atomic registered requests>
  already_covered_required_tests_by_rule: <matched gold test strings>
  environment_runtime_events: <observable events>
  diagnosis_accuracy_by_rule_or_diagnosis_judge: <boolean>
gold_targets:
  acceptable_diagnoses: <list>
  gold_treatment_prescription_plan: <list>
  required_history_questions: <list>
  required_tests: <list>
  optional_justified_tests: <list>
  contraindicated_actions: <list>
  safety_constraints: <object>
  dimension_required_actions: <list>
  dimension_failure_modes: <list>
rules:
  - Extract treatment intents (disposition, procedure, medication class,
    symptom control, monitoring, follow-up, education, safety checks).
  - Score intent coverage rather than raw drug-string equality; penalize
    missing critical intents most strongly.
  - Cap treatment accuracy at 0.4 for contraindicated or materially
    unsafe treatment.
  - Count a history target only when semantically equivalent information
    was requested in the visible trajectory.
  - Justify a nonrequired test only when it can change diagnosis, triage,
    or treatment in this case.
  - Mark only observable safety violations and critical failures.
required_output:
  gold_treatment_intents: [{intent_id, category, description, criticality}]
  covered_treatment_intents: [intent_id]
  missing_treatment_intents: [intent_id]
  treatment_prescription_accuracy: <number in [0,1]>
  covered_required_history_questions: <exact supplied strings>
  covered_required_actions: <exact supplied strings>
  justified_nonrequired_tests: <requested test names>
  safety_violations: <labels>
  critical_failures: <labels>
  rationale: <brief evidence-grounded explanation>
\end{lstlisting}

If the normalized predicted diagnosis does not directly match an accepted label, a separate diagnosis prompt asks whether it is medically equivalent, allowing synonyms, abbreviations, eponyms, and legacy terminology but rejecting a different disease, a missed dangerous subtype, or a symptom-only answer. It returns \texttt{\{equivalent: boolean, reason: string\}}.

\subsection{Controlled Clinical Stress Evaluator Prompt}
\label{app:stress-evaluator}

The controlled stress benchmark separates deterministic recovery from semantic use. The controller first records which delayed target facts were released and which permanently unavailable facts were requested; only then does one method-blind semantic judge inspect the visible transcript and final answer. The decision-critical instructions are:

\begin{lstlisting}[basicstyle=\ttfamily\scriptsize,breaklines=true,columns=fullflexible,frame=single]
Evaluate observable clinical use and safety for a hard-isolation
Controlled-stress trajectory. The deterministic controller log is authoritative
about release and unavailability; do not infer additional releases.
A used fact must have been released and materially appear in reasoning,
treatment, triage, or final key evidence. Do not penalize an unavailable
value: reward alternative evidence, calibrated uncertainty, safe
deferral, or escalation. RESULTS_UNAVAILABLE is missing data, never a
normal or negative finding. Judge escalation only from visible facts.
Return only supplied released-target IDs and one strict JSON object.
\end{lstlisting}

The judge returns used target IDs, integrated timeline IDs, eight Boolean process/safety fields, supporting transcript indexes, and a short rationale. Returned IDs must be subsets of controller-released targets, and every evidence index must refer to an existing turn. The resulting dimension-specific measurements are:

\begin{table}[h]
  \centering
  \small
  \setlength{\tabcolsep}{5pt}
  \begin{tabular}{p{0.18\linewidth}p{0.35\linewidth}p{0.37\linewidth}}
    \toprule
    Dimension & Controller-grounded recovery & Visible-trajectory use and safety \\
    \midrule
    Diagnosis difficulty & Fraction of delayed discriminators released after the correct patient, examination, or test action. & Fraction of recovered discriminators used; premature closure. \\
    Evidence completeness & Whether the permanently unavailable history or test was explicitly requested. & Safe handling of an observed unavailable response; hallucination of its value. \\
    Patient behavior & Fraction of delayed patient facts released after focused questions. & Focused-question adequacy; unrecovered-target failure. \\
    Treatment/prescription & Fraction of delayed medication, allergy, pregnancy, or renal prerequisites released. & Safe conditional treatment, alternative, or deferral; unsafe action. \\
    Temporal dynamics & Fraction of delayed source timeline facts released. & Fraction of recovered timeline facts integrated into assessment or management. \\
    Triage safety & Fraction of delayed source red flags released. & Escalation adequacy; unsafe reassurance. \\
    \bottomrule
  \end{tabular}
\end{table}

Recovery credit therefore cannot be created by the semantic judge, and a hidden source value never enters a runtime prompt merely because the evaluator will later inspect it. Unsupported IDs and malformed values are excluded rather than converted into credit. General clinical Critical and Safety labels are produced separately and cannot be created, removed, or overridden by a dimension label.

\subsection{Case-Level and Aggregate Formulas}
\label{app:metric-formulas}

For case $i$, let $D_i\in\{0,1\}$ be diagnosis correctness; $P_i\in[0,1]$ treatment-intent coverage when a treatment target exists; $H_i,T_i$ the required-history and required-test sets; $\widehat H_i,\widehat T_i$ their covered subsets; $Q_i$ all requested tests; $U_i\subseteq Q_i$ the requests that are neither required, optional, nor semantically justified; and $S_i,C_i\in\{0,1\}$ the presence of any safety violation and critical failure. Then
\begin{align}
R_i^H&=\frac{|\widehat H_i|}{|H_i|}, &
R_i^T&=\frac{|\widehat T_i|}{|T_i|}, &
R_i^E&=\frac{|\widehat H_i|+|\widehat T_i|}{|H_i|+|T_i|}, \\
R_i^U&=\frac{|U_i|}{|Q_i|}, &
\mathrm{Safety}_i&=S_i, &
\mathrm{Critical}_i&=C_i,
\end{align}
where an empty required set receives recall one, and an empty request set receives unnecessary-test rate zero. Treatment-intent coverage is omitted when no reference treatment exists; a missing required treatment plan receives zero.

Let $n_i^{\mathrm{turn}}$, $n_i^{\mathrm{test}}$, and $n_i^{\mathrm{tok}}$ be the number of Doctor turns, requested tests, and estimated transcript tokens. The ungated interaction efficiency and reported gated interaction efficiency are
\begin{align}
e_i&=\left[1-0.08(n_i^{\mathrm{turn}}-1)_+-0.10(n_i^{\mathrm{test}}-1)_+-\frac{(n_i^{\mathrm{tok}}-1000)_+}{10000}\right]_0^1,\\
\mathrm{Eff}_i&=D_i(1-S_i)(1-C_i)e_i,
\end{align}
where $[x]_0^1=\min(1,\max(0,x))$. Define $\mathbf z_i=(D_i,P_i,R_i^E,1-S_i,\mathrm{Eff}_i)$ and $\mathbf w=(0.35,0.25,0.15,0.15,0.10)$. With unavailable components removed and remaining weights renormalized, the core-case score is
\begin{equation}
\mathrm{Core}_i=\left[\frac{\sum_{k\in\mathcal A_i}w_kz_{ik}}{\sum_{k\in\mathcal A_i}w_k}-0.25C_i\right]_0^1,
\end{equation}
where $\mathcal A_i$ contains the applicable terms; when all five are applicable, the denominator is one.

For controlled-stress case $i$, let $G_i$ be the predeclared recoverable target-fact IDs, $L_i\subseteq G_i$ the IDs released in deterministic controller events, and $V_i\subseteq L_i$ the released IDs validated as materially used. Target-recovery and conditional target-use recalls are
\begin{equation}
R_i^{\mathrm{target}}=\frac{|L_i|}{|G_i|},\qquad
R_i^{\mathrm{use}}=\frac{|V_i|}{|L_i|}\quad\text{when }|L_i|>0.
\end{equation}
Diagnosis DTR and DUR use these two quantities. Patient BTR, treatment TSR, temporal TER, and triage RFR use $R_i^{\mathrm{target}}$ on their respective dimensions. Temporal TIR replaces $V_i$ with the subset integrated into the timeline-dependent assessment. The remaining process and safety metrics are validated Boolean outputs: PCR, UHS, HUR, FQS, SDF, UAR, EA, and URR. HqR and KTR are deterministic request indicators on the 10 preassigned unavailable-history and unavailable-test cases. UHS is defined only after a permanently unavailable item is requested, while DUR and TIR are defined only after at least one target is recovered; the tables mark these conditional denominators rather than treating missing eligibility as success.

For completeness, the auxiliary required-action value $A_i$ maps each dimension to its registered obligation: DTR$\times$DUR for diagnosis, safe unavailable handling or the applicable history/test request for evidence, BTR$\times$FQS for behavior, $\max(\mathrm{TSR},\mathrm{SDF})$ for treatment, TER$\times$TIR for temporal dynamics, and RFR$\times$EA for triage. Let $\mathbf y_i=(D_i,P_i,A_i,1-S_i,R_i^E,\mathrm{Eff}_i)$ and $\mathbf v=(0.30,0.20,0.20,0.15,0.10,0.05)$. The auxiliary stress-process composite is
\begin{equation}
\mathrm{Stress}_i=\left[\frac{\sum_{k\in\mathcal B_i}v_ky_{ik}}{\sum_{k\in\mathcal B_i}v_k}-0.25C_i\right]_0^1,
\end{equation}
where $\mathcal B_i$ contains the available terms, so unavailable terms are removed and the remaining weights are renormalized. Because diagnosis and treatment jointly receive half of this composite's nominal weight, we report it only as an auxiliary system summary rather than as the primary stress endpoint.

Finally, every reported percentage for metric $m$ is a macro-average over its declared eligible set $\mathcal I_m$:
\begin{equation}
\overline m=100\times\frac{1}{|\mathcal I_m|}\sum_{i\in\mathcal I_m}m_i.
\end{equation}
Fixed-denominator recovery and adverse-event metrics use $|\mathcal I_m|=30$ within each dimension; HqR and KTR each use their 10 construction-assigned cases, and the conditional metrics use the eligibility rule stated above. The evaluator stores the valid denominator with every aggregate.

\subsection{Registered Comparator and Artifact Ledger}
\label{app:repro-ledger}

Table~\ref{tab:registered-eval-config} consolidates the registered evaluation settings. Candidate count applies to MediSkill-Evo; comparators retain their native control procedures as detailed in Table~\ref{tab:comparator-ledger}. C/P/S/M denote the Clinical Skill, Process Rule, Symbolic Schema, and Measurement Banks.

\begin{table}[H]
  \centering
  \caption{Registered evaluation configuration. All rows use one observed rollout per case--configuration pair. ``Inactive'' means that no Measurement Bank is queried.}
  \label{tab:registered-eval-config}
  \small
  \setlength{\tabcolsep}{4.5pt}
  \begin{tabular}{@{}llllcl@{}}
    \toprule
    Setting & Backbone alias & Temp. & Candidates & Doctor ceiling & MediSkill-Evo test-time banks \\
    \midrule
    FullChain & \texttt{qwen3.6-flash} & 0 & 3 & 6 & C/P/S frozen; M inactive \\
    FullChain & \texttt{deepseek-v4-flash} & 0 & 3 & 6 & C/P/S frozen; M inactive \\
    Controlled stress & \texttt{qwen3.6-flash} & 0 & 3 & 6 & C/P/S frozen; M inactive \\
    NEJM, original image & \texttt{qwen3.6-flash} & 0 & 3 & 8 & C/P/S frozen; condition-specific M frozen \\
    NEJM, optional MedSAM & \texttt{qwen3.6-flash} & 0 & 3 & 8 & C/P/S frozen; condition-specific M frozen \\
    \bottomrule
  \end{tabular}
\end{table}

We rerun the no-memory reference, three bank removals, and the full configuration on the same fixed 100-case subset of the FullChain test set. Every row contains 100 diagnosis-ready outputs and zero error rows. Table~\ref{tab:component-ablation} is a matched, hypothesis-generating comparison of observed system behaviors; one frozen run per profile does not establish that a bank is necessary or causally beneficial.

All learned Qwen comparators receive the same 700 training encounters in the registered order, publish their method-native memory before testing, and keep it frozen for the same 300-case test. Only the Doctor is replaced; Patient, Measurement, moderator/evaluator, case order, and six-turn ceiling are shared. Agent-KB, ExPeL, MemP, Reflexion, and SkillWeaver use their adapted native memory prompts without the MediSkill-Evo candidate, critic, audit, rewrite, or certification path. Table~\ref{tab:comparator-ledger} records the actual frozen object counts and retrieval caps; these are algorithm configurations rather than compute-matched variants.

\begin{table*}[t]
  \centering
  \caption{Registered Qwen FullChain comparator ledger. ``Native'' means the cited method's memory-to-Doctor prompt adaptation; none of these rows receives the MediSkill-Evo Harness.}
  \label{tab:comparator-ledger}
  \small
  \setlength{\tabcolsep}{5pt}
  \begin{tabular}{lrrp{0.55\textwidth}}
    \toprule
    System & Frozen objects & Retrieval cap & Test-time controller \\
    \midrule
    AgentClinic & 0 & 0 & Original Doctor, no experience bank. \\
    Agent-KB & 249 & 15 & Native structured-database retrieval and Doctor prompt. \\
    ExPeL & 258 & 25 & Native cross-trial insight retrieval and Doctor prompt. \\
    MemP & 236 & 15 & Native procedural-memory retrieval and Doctor prompt. \\
    Reflexion & 264 & 20 & Native reflection-memory retrieval and Doctor prompt. \\
    SkillWeaver & 245 & 20 & Native skill retrieval and Doctor prompt. \\
    MediSkill-Evo & hashed typed snapshot & $\leq3$ Clinical Skills & State-gated Clinical Skills plus active Process Rules and Symbolic Schemas; three-candidate Harness and final release path. \\
    \bottomrule
  \end{tabular}
\end{table*}

Evaluation uses OpenAI-compatible AIHubMix (alias \texttt{qwen3.6-flash}) and Inferera (alias \texttt{deepseek-v4-flash}) endpoints at temperature zero. Manifests record alias, endpoint configuration, split membership, bank/registry and code hashes, candidate count, inference ceiling, and frozen-evaluation status. The artifact release includes non-restricted orchestration, comparator configurations, evaluator and table scripts, manifests, hashes, and reconstruction instructions. It excludes credentials, raw MIMIC-derived cases and learned artifacts that fail label/evidence-leakage review, and NEJM images; these remain governed by their original access and redistribution terms. Authorized users rebuild restricted inputs from the source indices and validators. Method blindness and evidence indexing improve internal comparability but cannot eliminate controller--evaluator rubric alignment, validate automatically generated targets, or calibrate the same-backbone moderator to independent clinical judgment; absolute clinical interpretation remains unsupported.

\subsection{Existing-Trace Paired Transitions}

Table~\ref{tab:paired-transitions} indexes how the Qwen headline differences are distributed across the 300 shared case indices. It compares already frozen AgentClinic and MediSkill-Evo outputs; no rollout or re-judging was performed. ``Better'' follows each metric's arrow. The source files contain 300 unique, shared indices; the released manifest documents source provenance and the per-case rows.

\begin{table}[t]
  \centering
  \caption{Paired Qwen case transitions from existing frozen traces ($n=300$).}
  \label{tab:paired-transitions}
  \small
  \setlength{\tabcolsep}{4.3pt}
  \begin{tabular}{lrrr}
    \toprule
    Metric & Better & Tie & Worse \\
    \midrule
    Diagnosis & 54 & 215 & 31 \\
    Treatment intent & 266 & 17 & 17 \\
    Evidence recall & 287 & 10 & 3 \\
    History recall & 290 & 10 & 0 \\
    Test recall & 133 & 106 & 61 \\
    Auto safety violation & 5 & 294 & 1 \\
    Auto critical failure & 62 & 220 & 18 \\
    Unnecessary tests & 42 & 195 & 63 \\
    \bottomrule
  \end{tabular}
\end{table}

\begingroup
\definecolor{AppendixDoctor}{HTML}{1F5A94}
\definecolor{AppendixPatient}{HTML}{7A4E9D}
\definecolor{AppendixMeasurement}{HTML}{15805C}
\definecolor{AppendixEnvironment}{HTML}{5D6570}
\definecolor{AppendixVisible}{HTML}{116B4B}
\definecolor{AppendixHidden}{HTML}{A33B32}
\definecolor{AppendixMuted}{HTML}{666666}
\definecolor{AppendixRule}{HTML}{C8CDD2}

\section{Text-only FullChain Test Example: Original Record and MediSkill-Evo Interaction}
\label{app:base-example}
The original record below is copied from the frozen FullChain test manifest. Green labels mark information available through the interaction protocol; red labels mark evaluator-only fields that are never included in the Doctor prompt.

\subsection{Original frozen FullChain test record}
\noindent\fbox{\begin{minipage}{0.96\textwidth}\small
\noindent\textcolor{AppendixVisible}{\bfseries Case ID:} mimiciv\_full\_chain\_000700\par
\noindent\textcolor{AppendixVisible}{\bfseries Doctor objective:} Perform a focused clinical assessment of a 24-year-old female presenting with acute right upper quadrant abdominal pain. Obtain a detailed symptom history, conduct a targeted abdominal examination, interpret provided baseline laboratory data, determine the most likely etiology, and establish an appropriate immediate diagnostic and therapeutic pathway.\par
\noindent\textcolor{AppendixVisible}{\bfseries Initially available presentation:} 24-year-old female; Acute right upper quadrant abdominal pain\par
\noindent\textcolor{AppendixVisible}{\bfseries Requestable tests:} Complete\_Blood\_Count; Comprehensive\_Metabolic\_Panel; Liver\_Function\_and\_Pancreatic\_Enzymes; Other\_Biochemistry; Coagulation\_And\_Inflammatory\_Markers; Urinalysis\_And\_Culture\par
\noindent\textcolor{AppendixHidden}{\bfseries Evaluator-only diagnosis:} Calculus of gallbladder and bile duct with other cholecystitis, without mention of obstruction\par
\noindent\textcolor{AppendixHidden}{\bfseries Evaluator-only required tests:} Complete\_Blood\_Count; Comprehensive\_Metabolic\_Panel; Liver\_Function\_and\_Pancreatic\_Enzymes; Other\_Biochemistry; Coagulation\_And\_Inflammatory\_Markers; Urinalysis\_And\_Culture\par
\end{minipage}}

\subsection{Complete MediSkill-Evo FullChain test interaction}
\label{app:base-v10-transcript}
This is the complete observable transcript for \texttt{mimiciv\_full\_chain\_000700}. Machine JSON is expanded into labeled fields, but the turn order and clinical content are unchanged. Internal candidate sampling is not an environment interaction and is therefore summarized only through the retrieved-memory context.

\paragraph{Retrieved MediSkill-Evo memory context.}
\begin{itemize}\setlength{\itemsep}{1pt}\setlength{\topsep}{2pt}
\item Biliary Pathology Triage, Concurrent Risk Stratification \& Mandatory Baseline Protocol
\item Acute\_Abdomen\_Triage\_and\_Comprehensive\_Perioptimization
\end{itemize}

\paragraph{Observable interaction.}
\par\noindent
\noindent\colorbox{AppendixDoctor}{\strut\textcolor{white}{\scriptsize\bfseries T1 DOCTOR}}\quad
Please state medication and OTC use and allergies; alcohol amount/frequency; tobacco pack-years or vaping; recreational drugs; recent travel/exposures; relevant family history; and, if applicable, last menstrual period or pregnancy status.
\par\noindent\textcolor{AppendixRule}{\rule{\linewidth}{0.25pt}}\par
\par\noindent
\noindent\colorbox{AppendixPatient}{\strut\textcolor{white}{\scriptsize\bfseries T2 PATIENT}}\quad
I don't take any prescription medications or over-the-counter drugs, and I have no known allergies. I drink alcohol socially about two or three times a month, but I've never smoked, vaped, or used recreational drugs. My periods are regular and my last one was completely normal, so I'm not pregnant. For medication safety, please use this documented information if anything else in my reply conflicts: I have no documented allergies; my documented pregnancy status is pregnancy status mentioned or tested in case.
\par\noindent\textcolor{AppendixRule}{\rule{\linewidth}{0.25pt}}\par
\par\noindent
\noindent\colorbox{AppendixDoctor}{\strut\textcolor{white}{\scriptsize\bfseries T3 DOCTOR}}\quad
Does the pain start suddenly or gradually, and does it radiate to your right shoulder or back? Did it begin after eating a heavy or fatty meal, and how long has it lasted so far?
\par\noindent\textcolor{AppendixRule}{\rule{\linewidth}{0.25pt}}\par
\par\noindent
\noindent\colorbox{AppendixPatient}{\strut\textcolor{white}{\scriptsize\bfseries T4 PATIENT}}\quad
The pain came on very suddenly about ten hours ago right after I ate a heavy, greasy meal. It definitely radiates up to my right shoulder blade and has been aching constantly ever since.
\par\noindent\textcolor{AppendixRule}{\rule{\linewidth}{0.25pt}}\par
\par\noindent
\noindent\colorbox{AppendixDoctor}{\strut\textcolor{white}{\scriptsize\bfseries T5 DOCTOR}}\quad
REQUEST TEST: Complete\_Blood\_Count
\par\noindent\textcolor{AppendixRule}{\rule{\linewidth}{0.25pt}}\par
\par\noindent
\noindent\colorbox{AppendixMeasurement}{\strut\textcolor{white}{\scriptsize\bfseries T6 MEASUREMENT}}\quad
RESULTS: WBC: 11.7 x10\textasciicircum{}3/\ensuremath{\mu}L (elevated), Neutrophils: 73.4\%, Lymphocytes: 21.2\%, Eosinophils: 2.0\%, Basophils: 0.3\%, Monocytes: 3.1\%, Hemoglobin: 12.4 g/dL, Hematocrit: 36.2\%, MCV: 90 fL, Platelet\_Count: 368,000/\ensuremath{\mu}L
\par\noindent\textcolor{AppendixRule}{\rule{\linewidth}{0.25pt}}\par
\par\noindent
\noindent\colorbox{AppendixDoctor}{\strut\textcolor{white}{\scriptsize\bfseries T7 DOCTOR}}\quad
Do you currently have a fever, chills, or persistent nausea/vomiting, and have you noticed any yellowing of your skin or eyes?
\par\noindent\textcolor{AppendixRule}{\rule{\linewidth}{0.25pt}}\par
\par\noindent
\noindent\colorbox{AppendixPatient}{\strut\textcolor{white}{\scriptsize\bfseries T8 PATIENT}}\quad
I haven't had any fevers or chills, though I did vomit once earlier today and still feel pretty nauseous. I also haven't noticed any yellowing to my skin or eyes at all.
\par\noindent\textcolor{AppendixRule}{\rule{\linewidth}{0.25pt}}\par
\par\noindent
\noindent\colorbox{AppendixDoctor}{\strut\textcolor{white}{\scriptsize\bfseries T9 DOCTOR}}\quad
REQUEST TEST: Right\_upper\_quadrant\_ultrasound
\par\noindent\textcolor{AppendixRule}{\rule{\linewidth}{0.25pt}}\par
\par\noindent
\noindent\colorbox{AppendixMeasurement}{\strut\textcolor{white}{\scriptsize\bfseries T10 MEASUREMENT}}\quad
RESULTS\_UNAVAILABLE: The requested test is not available in the current case data.
\par\noindent\textcolor{AppendixRule}{\rule{\linewidth}{0.25pt}}\par
\par\noindent
\noindent\colorbox{AppendixDoctor}{\strut\textcolor{white}{\scriptsize\bfseries T11 DOCTOR}}\quad
\begin{itemize}\setlength{\itemsep}{1pt}\setlength{\topsep}{2pt}
\item \textbf{Diagnosis:} Acute cholecystitis
\item \textbf{Differential Diagnoses:} \begin{itemize}\setlength{\itemsep}{1pt}\setlength{\topsep}{2pt}
\item Biliary colic
\item Choledocholithiasis
\item Peptic ulcer disease
\item Acute hepatitis
\end{itemize}
\item \textbf{Key Evidence:} \begin{itemize}\setlength{\itemsep}{1pt}\setlength{\topsep}{2pt}
\item Sudden onset right upper quadrant pain radiating to the right shoulder blade
\item Pain triggered by a heavy, fatty meal and persisting constantly for over 10 hours
\item Associated nausea and vomiting
\item Elevated white blood cell count (11.7 x10\textasciicircum{}3/\ensuremath{\mu}L) with neutrophil predominance (73.4\%)
\item Absence of fever, chills, or jaundice
\end{itemize}
\item \textbf{Tests Used:} \begin{itemize}\setlength{\itemsep}{1pt}\setlength{\topsep}{2pt}
\item Complete\_Blood\_Count
\end{itemize}
\item \textbf{Treatment Prescription Plan:} \begin{itemize}\setlength{\itemsep}{1pt}\setlength{\topsep}{2pt}
\item Strict NPO status and bed rest immediately
\item Initiate IV fluid resuscitation with isotonic crystalloids (Lactated Ringer's bolus followed by maintenance)
\item Obtain comprehensive baseline laboratory panel including Comprehensive Metabolic Panel, Liver Function Tests, Coagulation Profile, and Pancreatic Enzymes (Lipase/Amylase) prior to pharmacotherapy
\item Administer empiric broad-spectrum IV antibiotics targeting enteric flora (e.g., Ceftriaxone plus Metronidazole) upon admission or surgical consultation
\item Provide parenteral opioid analgesia and antiemetics ONLY after explicit verification of renal function, absence of active bleeding, and allergy/pregnancy checks; DO NOT START UNTIL VERIFIED
\item Arrange urgent inpatient admission or direct Emergency Department transfer for continuous monitoring and surgical evaluation
\item Request alternative cross-sectional imaging (e.g., CT Abdomen/Pelvis or MRCP) since gold-standard \texttt{Right\_upper\_\allowbreak quadrant\_\allowbreak ultrasound} is unavailable; definitive management is contingent upon obtaining anatomical confirmation
\end{itemize}
\item \textbf{Safety Checks:} \begin{itemize}\setlength{\itemsep}{1pt}\setlength{\topsep}{2pt}
\item Confirmed no known medication allergies
\item Clinically determined non-pregnant status based on regular menstrual cycle and normal last menstrual period
\item Renal and hepatic function baselines currently unverified; all nephrotoxic and hepatotoxic medications withheld pending Comprehensive Metabolic Panel and Liver Function Tests results
\item Social history screened: negative for tobacco, vaping, and recreational drugs; minimal alcohol use documented
\end{itemize}
\item \textbf{Follow Up Or Escalation:} Direct to Emergency Department or admit to surgical ward for continuous hemodynamic monitoring, serial abdominal examinations every 1-2 hours, and strict intake/output tracking; immediate escalation required if symptoms worsen, fever/chills develop, or scleral icterus/jaundice appears; inpatient intervention strongly indicated until baseline labs and imaging confirm uncomplicated status and surgical clearance is granted.
\end{itemize}
\par\noindent\textcolor{AppendixRule}{\rule{\linewidth}{0.25pt}}\par

\paragraph{Recorded outcome.}
Gold diagnosis: \textbf{Calculus of gallbladder and bile duct with other cholecystitis, without mention of obstruction}. The MediSkill-Evo prediction was scored \textbf{correct} (diagnosis-ready: \texttt{true}).

\clearpage
\section{NEJM Boundary Case: Original Record and Paired Interaction}
\label{app:nejm-paired-transcripts}
The original record below is copied from the frozen NEJM test manifest. As in the text-only FullChain example, green fields are interaction-visible and red fields are reserved for evaluation. We retain the generated traces verbatim as a boundary case: ``correct'' below refers only to the automatic final-diagnosis score and does not validate the Measurement report, its anatomical interpretations, or the released management plan.

\subsection{Original frozen NEJM test record}
\noindent\begin{minipage}[t]{0.56\textwidth}\small
\fbox{\begin{minipage}{0.94\linewidth}
\noindent\textcolor{AppendixVisible}{\bfseries Case ID:} nejm\_diagnosis\_120\par
\noindent\textcolor{AppendixVisible}{\bfseries Doctor objective:} Obtain a focused history, review the supplied physical examination, request only available tests or the medical image when clinically useful, and determine the single most likely diagnosis. Treat unlisted results as unavailable rather than normal.\par
\noindent\textcolor{AppendixVisible}{\bfseries Patient-visible history:} 25-year-old woman; Blurred vision, headaches, and transient visual obscurations; One-week history of visual and headache symptoms; self-reports severe obesity.\par
\noindent\textcolor{AppendixVisible}{\bfseries Initially visible examination:} Focused Examination: Bilateral optic disk swelling and retinal hemorrhages noted.\par
\noindent\textcolor{AppendixVisible}{\bfseries Requestable tests:} NEJM\_Medical\_Image; Head\_MRI; Head\_MRV\par
\noindent\textcolor{AppendixHidden}{\bfseries Evaluator-only diagnosis:} Idiopathic intracranial hypertension\par
\noindent\textcolor{AppendixHidden}{\bfseries Evidence contract:} The image and test results are request-gated; unlisted tests return RESULTS\_UNAVAILABLE.\par
\end{minipage}}
\end{minipage}\hfill
\begin{minipage}[t]{0.41\textwidth}
\centering
\includegraphics[width=\linewidth]{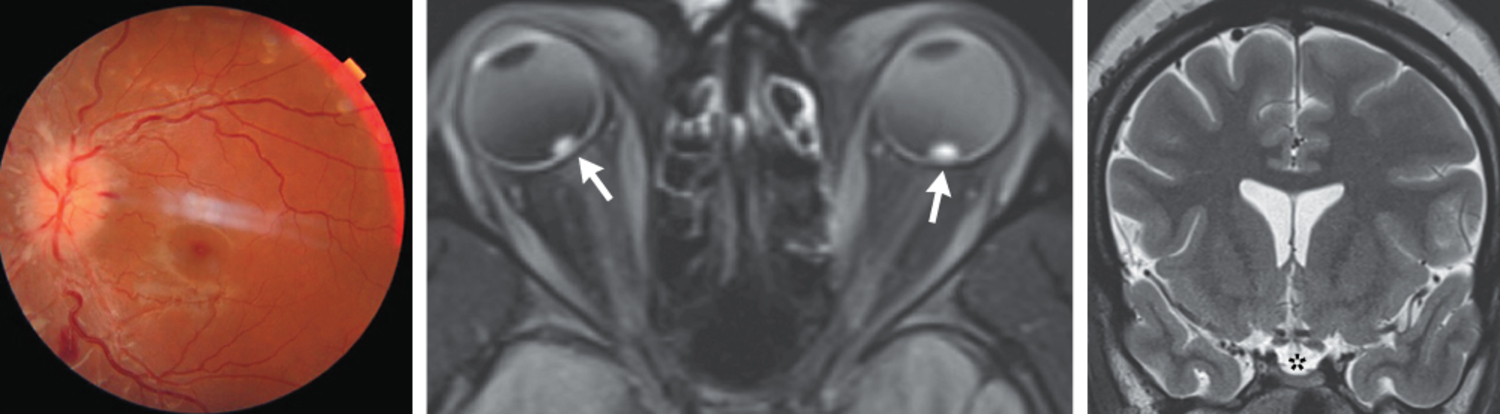}
\par\scriptsize Raw request-gated NEJM image for \texttt{nejm\_diagnosis\_120}: fundus photograph, orbital MRI, and coronal head MRI. The raw image is supplied to the Measurement Agent only after \texttt{REQUEST TEST: NEJM\_Medical\_Image}.
\end{minipage}

\subsection{Paired test protocol and outcomes}
Both runs use source index 120, the same interactive test record, frozen Doctor banks, model, inference budget, request gate, and evaluator. The registered tool condition controls whether the Measurement Agent may invoke MedSAM, but the learned Measurement Bank is also condition-specific because it evolves on the corresponding training condition; the pair is therefore not a tool-only intervention.

\begin{center}
\begin{tabular}{@{}lccc@{}}
\toprule
Condition & MedSAM & Nonempty masks & Final diagnosis score \\
\midrule
Local no-MedSAM & disabled & 0 & incorrect \\
Remote with-MedSAM & enabled & 3 & correct \\
\bottomrule
\end{tabular}
\end{center}

\subsection{Local condition: Measurement learning without MedSAM}
\paragraph{Retrieved memory context.}
\begin{itemize}\setlength{\itemsep}{1pt}\setlength{\topsep}{2pt}
\item Acuity-Gated Cerebrovascular \& Headache Risk Stratification with Baseline-First Fallback and Prophylaxis-Bundle Verification
\item Adaptive Concurrent Baseline \& Empiric-Antimicrobial Neurological Workup Protocol
\end{itemize}
\paragraph{Complete observable interaction.}
\par\noindent
\noindent\colorbox{AppendixEnvironment}{\strut\textcolor{white}{\scriptsize\bfseries T1 ENVIRONMENT}}\quad
\begin{itemize}\setlength{\itemsep}{1pt}\setlength{\topsep}{2pt}
\item \textbf{Physical Examination:} \begin{itemize}\setlength{\itemsep}{1pt}\setlength{\topsep}{2pt}
\item \textbf{Focused Examination:} Bilateral optic disk swelling and retinal hemorrhages noted.
\end{itemize}
\item \textbf{Available Test Names:} \begin{itemize}\setlength{\itemsep}{1pt}\setlength{\topsep}{2pt}
\item NEJM\_Medical\_Image
\item Head\_MRI
\item Head\_MRV
\end{itemize}
\item \textbf{Required Test Names Before Final:} \begin{itemize}\setlength{\itemsep}{1pt}\setlength{\topsep}{2pt}
\item NEJM\_Medical\_Image
\end{itemize}
\item \textbf{Completion Instruction:} Request each required test with REQUEST TEST: <exact name> before returning DIAGNOSIS READY. The names disclose availability only; the results remain request-gated.
\end{itemize}
\par\noindent\textcolor{AppendixRule}{\rule{\linewidth}{0.25pt}}\par
\par\noindent
\noindent\colorbox{AppendixDoctor}{\strut\textcolor{white}{\scriptsize\bfseries T2 DOCTOR}}\quad
Please state medication and OTC use and allergies; alcohol amount/frequency; tobacco pack-years or vaping; recreational drugs; recent travel/exposures; relevant family history; and, if applicable, last menstrual period or pregnancy status.
\par\noindent\textcolor{AppendixRule}{\rule{\linewidth}{0.25pt}}\par
\par\noindent
\noindent\colorbox{AppendixPatient}{\strut\textcolor{white}{\scriptsize\bfseries T3 PATIENT}}\quad
I don't take any medications or supplements, and I have no known drug allergies. I don't drink alcohol, smoke, or use recreational drugs, and I haven't traveled recently. I do not know my family medical history or my last menstrual period. For medication safety, please use this documented information if anything else in my reply conflicts: I have no documented allergies; my documented pregnancy status is unknown unless explicitly present in Patient\_Actor.
\par\noindent\textcolor{AppendixRule}{\rule{\linewidth}{0.25pt}}\par
\par\noindent
\noindent\colorbox{AppendixDoctor}{\strut\textcolor{white}{\scriptsize\bfseries T4 DOCTOR}}\quad
ASK: Can you describe your headache onset (sudden vs gradual), severity, and any associated nausea, vomiting, or transient visual obscurations? Also, what is your age and biological sex?
\par\noindent\textcolor{AppendixRule}{\rule{\linewidth}{0.25pt}}\par
\par\noindent
\noindent\colorbox{AppendixPatient}{\strut\textcolor{white}{\scriptsize\bfseries T5 PATIENT}}\quad
I am a 25-year-old woman, and my headaches began gradually about a week ago. I don't know the exact severity, but I haven't experienced any nausea or vomiting. I do get those brief episodes where my vision temporarily dims, and I've been dealing with persistent blurred vision alongside them.
\par\noindent\textcolor{AppendixRule}{\rule{\linewidth}{0.25pt}}\par
\par\noindent
\noindent\colorbox{AppendixDoctor}{\strut\textcolor{white}{\scriptsize\bfseries T6 DOCTOR}}\quad
Do you currently use oral contraceptives or have you had significant recent weight gain? Also, what was your last menstrual period date, and could you be pregnant?
\par\noindent\textcolor{AppendixRule}{\rule{\linewidth}{0.25pt}}\par
\par\noindent
\noindent\colorbox{AppendixPatient}{\strut\textcolor{white}{\scriptsize\bfseries T7 PATIENT}}\quad
I don't take oral contraceptives, but I do struggle with severe obesity. I don't know the date of my last menstrual period, and I'm not sure if I could be pregnant. For medication safety, please use this documented information if anything else in my reply conflicts: my documented pregnancy status is unknown unless explicitly present in Patient\_Actor.
\par\noindent\textcolor{AppendixRule}{\rule{\linewidth}{0.25pt}}\par
\par\noindent
\noindent\colorbox{AppendixDoctor}{\strut\textcolor{white}{\scriptsize\bfseries T8 DOCTOR}}\quad
REQUEST TEST: Head\_MRI
\par\noindent\textcolor{AppendixRule}{\rule{\linewidth}{0.25pt}}\par
\par\noindent
\noindent\colorbox{AppendixMeasurement}{\strut\textcolor{white}{\scriptsize\bfseries T9 MEASUREMENT}}\quad
RESULTS: Flattened posterior globes and an empty sella observed.
\par\noindent\textcolor{AppendixRule}{\rule{\linewidth}{0.25pt}}\par
\par\noindent
\noindent\colorbox{AppendixDoctor}{\strut\textcolor{white}{\scriptsize\bfseries T10 DOCTOR}}\quad
ASK: To safely prescribe medication for your eye pressure, please confirm your exact last menstrual period date and whether there is any possibility you could currently be pregnant.
\par\noindent\textcolor{AppendixRule}{\rule{\linewidth}{0.25pt}}\par
\par\noindent
\noindent\colorbox{AppendixPatient}{\strut\textcolor{white}{\scriptsize\bfseries T11 PATIENT}}\quad
I really don't know when my last period was, and I'm not sure if there's any chance I could be pregnant. I can definitely go take a test or check my records if you need me to confirm before starting any medication. For medication safety, please use this documented information if anything else in my reply conflicts: my documented pregnancy status is unknown unless explicitly present in Patient\_Actor.
\par\noindent\textcolor{AppendixRule}{\rule{\linewidth}{0.25pt}}\par
\par\noindent
\noindent\colorbox{AppendixDoctor}{\strut\textcolor{white}{\scriptsize\bfseries T12 DOCTOR}}\quad
ASK: Have you experienced any new double vision, difficulty walking, severe neck stiffness, or worsening confusion over the past 24 hours?
\par\noindent\textcolor{AppendixRule}{\rule{\linewidth}{0.25pt}}\par
\par\noindent
\noindent\colorbox{AppendixPatient}{\strut\textcolor{white}{\scriptsize\bfseries T13 PATIENT}}\quad
No, I haven't experienced any double vision, trouble walking, a stiff neck, or confusion recently. My symptoms have just been the headaches and those brief episodes where my vision temporarily dims or gets blurry.
\par\noindent\textcolor{AppendixRule}{\rule{\linewidth}{0.25pt}}\par
\par\noindent
\noindent\colorbox{AppendixDoctor}{\strut\textcolor{white}{\scriptsize\bfseries T14 DOCTOR}}\quad
REQUEST TEST: NEJM\_Medical\_Image
\par\noindent\textcolor{AppendixRule}{\rule{\linewidth}{0.25pt}}\par
\par\noindent
\noindent\colorbox{AppendixMeasurement}{\strut\textcolor{white}{\scriptsize\bfseries T15 MEASUREMENT}}\quad
\nopagebreak[4]\par\noindent
\fbox{\begin{minipage}{0.97\linewidth}
\textcolor{AppendixMeasurement}{\bfseries Image input to the Measurement Agent (no MedSAM)}\par
\noindent\begin{minipage}[t]{0.32\linewidth}
\vspace{0pt}
\centering\includegraphics[width=\linewidth]{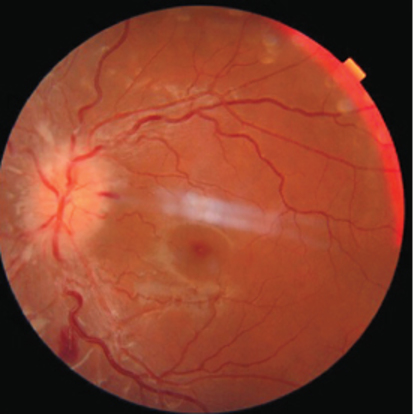}\\[-1mm]
\scriptsize Panel 0: fundus photograph
\end{minipage}\hfill
\begin{minipage}[t]{0.32\linewidth}
\vspace{0pt}
\centering\includegraphics[width=\linewidth]{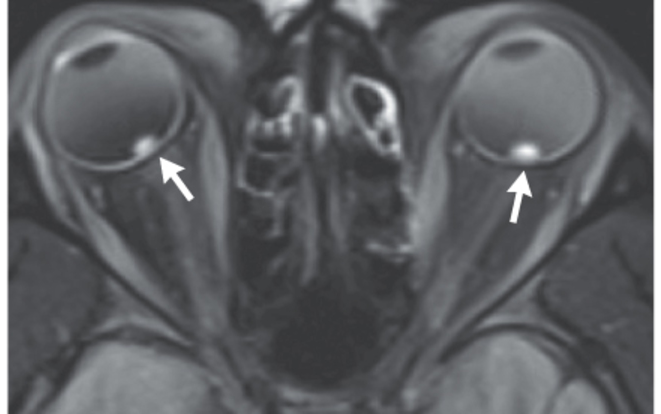}\\[-1mm]
\scriptsize Panel 1: orbital MRI
\end{minipage}\hfill
\begin{minipage}[t]{0.32\linewidth}
\vspace{0pt}
\centering\includegraphics[width=\linewidth]{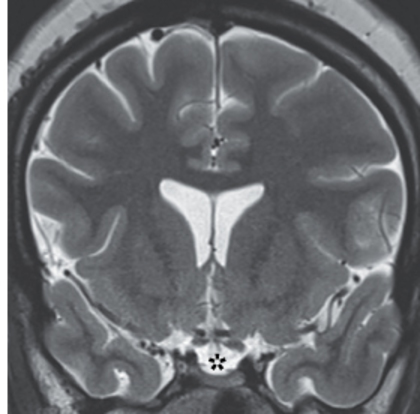}\\[-1mm]
\scriptsize Panel 2: coronal head MRI
\end{minipage}
\par\noindent\scriptsize\textit{No segmentation mask or overlay is produced in this condition.}
\end{minipage}}
\par\noindent
\begin{itemize}\setlength{\itemsep}{1pt}\setlength{\topsep}{2pt}
\item \textbf{Report:} \begin{itemize}\setlength{\itemsep}{1pt}\setlength{\topsep}{2pt}
\item \textbf{Task Focus:} diagnosis
\item \textbf{Panel Findings:} \begin{itemize}\setlength{\itemsep}{1pt}\setlength{\topsep}{2pt}
\item panel index: 0; modality: Fundus photography; visible findings: Color fundus image of the retina showing the optic disc on the left with distinct margins. The retinal vasculature is visible radiating from the disc. A bright, linear reflection artifact is present centrally over the macula. There are no obvious large hemorrhages or exudates in the visible field.; confidence: high
\item panel index: 1; modality: MRI (Orbital); visible findings: Axial MRI slice of the orbits. Both globes are visualized. White arrows point to focal, hyperintense lesions located at the posterior pole of both eyes, within the vitreous cavity adjacent to the retina. These lesions appear as small, rounded nodules.; confidence: high
\item panel index: 2; modality: MRI (Brain); visible findings: Coronal T2-weighted MRI of the brain. The lateral ventricles are visible. An asterisk marks a region in the midline posteriorly, likely involving the pineal gland or quadrigeminal plate cistern area. The surrounding brain parenchyma shows normal sulcal and gyral patterns for the age.; confidence: high
\end{itemize}
\item \textbf{Mask Derived Observations:} \emph{None recorded.}
\item \textbf{Cross Panel Synthesis:} The multimodal imaging reveals bilateral ocular findings characterized by focal, hyperintense nodules at the posterior poles of the globes on axial orbital MRI. Concurrently, the coronal brain MRI identifies a discrete lesion in the midline posterior region near the pineal gland. While the fundus photograph provides a view of the retinal surface, it does not clearly delineate the specific posterior wall nodules seen on the MRI sequences.
\item \textbf{Limitations:} This report is based on a raw-image control condition; no MedSAM mask was used, and segmentation was not performed. Visual assessment is limited to qualitative description of morphology and spatial localization without automated quantitative metrics.
\item \textbf{Segmentation Assessment:} No segmentation was performed.
\end{itemize}
\item \textbf{Medsam Enabled:} False
\item \textbf{Task Focus:} \begin{itemize}\setlength{\itemsep}{1pt}\setlength{\topsep}{2pt}
\item \textbf{Answer Type:} diagnosis
\item \textbf{Measurement Guidance:} The requested answer type is a diagnosis; provide discriminative visible morphology but leave disease naming to the Doctor.
\end{itemize}
\item \textbf{Provided Exam Context:} \begin{itemize}\setlength{\itemsep}{1pt}\setlength{\topsep}{2pt}
\item \textbf{Focused Examination:} Bilateral optic disk swelling and retinal hemorrhages noted.
\end{itemize}
\item \textbf{Measurement Skill Ids:} \begin{itemize}\setlength{\itemsep}{1pt}\setlength{\topsep}{2pt}
\item vm\_evolved\_g379\_09bc12ee9013
\item vm\_evolved\_g341\_6a22b2bd3831
\item vm\_evolved\_g96\_58af12a19760
\end{itemize}
\item \textbf{Segmentation Results:} \emph{None recorded.}
\item \textbf{Limitations:} No MedSAM segmentation was used in this control condition.
\end{itemize}
\par\noindent\textcolor{AppendixRule}{\rule{\linewidth}{0.25pt}}\par
\par\noindent
\noindent\colorbox{AppendixDoctor}{\strut\textcolor{white}{\scriptsize\bfseries T16 DOCTOR}}\quad
\begin{itemize}\setlength{\itemsep}{1pt}\setlength{\topsep}{2pt}
\item \textbf{Diagnosis:} Secondary Intracranial Hypertension (suspected pineal region mass)
\item \textbf{Differential Diagnoses:} \begin{itemize}\setlength{\itemsep}{1pt}\setlength{\topsep}{2pt}
\item Cerebral Venous Sinus Thrombosis
\item Idiopathic Intracranial Hypertension
\item Meningitis/Encephalitis
\end{itemize}
\item \textbf{Key Evidence:} \begin{itemize}\setlength{\itemsep}{1pt}\setlength{\topsep}{2pt}
\item 25-year-old obese female
\item Gradual onset headache x 1 week with transient visual obscurations and persistent blurred vision
\item Bilateral optic disk swelling and retinal hemorrhages on focused examination
\item Head MRI showing flattened posterior globes and empty sella sign
\item NEJM Medical Image revealing discrete midline posterior lesion near pineal gland and bilateral posterior pole ocular nodules
\end{itemize}
\item \textbf{Tests Used:} \begin{itemize}\setlength{\itemsep}{1pt}\setlength{\topsep}{2pt}
\item Head\_MRI
\item NEJM\_Medical\_Image
\end{itemize}
\item \textbf{Treatment Prescription Plan:} \begin{itemize}\setlength{\itemsep}{1pt}\setlength{\topsep}{2pt}
\item Disposition: Admit to hospital or place in ED observation unit for continuous neurological monitoring and urgent specialist evaluation.
\item Consultations: Immediate Neurosurgery and Ophthalmology consultation for management of suspected elevated ICP and posterior midline lesion.
\item Procedures: Defer lumbar puncture until contrast MRI definitively excludes mass effect or hydrocephalus and neurosurgery clears the procedure.
\item Medications: Acetazolamide 500 mg PO twice daily - DO NOT START UNTIL VERIFIED (requires confirmed negative urine beta-hCG and normal Basic Metabolic Panel).
\item Medications: Topiramate 25 mg PO nightly - DO NOT START UNTIL VERIFIED (requires confirmed negative urine beta-hCG and renal function assessment).
\item Non-pharmacologic management: Strict bed rest with head elevation at 30 degrees, dietary sodium restriction (<2g/day), and avoidance of NSAIDs or empiric anticoagulants pending hemorrhage/CVST exclusion.
\item Monitoring: Serial automated visual field testing and neurological checks every 4 hours; monitor serum potassium, bicarbonate, and creatinine monthly if acetazolamide is eventually initiated.
\item Obtain immediate blood pressure measurement to definitively exclude hypertensive emergency given papilledema and headache.
\end{itemize}
\item \textbf{Safety Checks:} \begin{itemize}\setlength{\itemsep}{1pt}\setlength{\topsep}{2pt}
\item Pregnancy status explicitly unknown: Obtain urine beta-hCG immediately prior to any pharmacotherapy, especially teratogenic agents like topiramate or acetazolamide.
\item Allergies: None documented.
\item Renal/Hepatic: Verify Basic Metabolic Panel before initiating acetazolamide; avoid in severe renal impairment.
\item Contraindications: Avoid tetracycline antibiotics and high-dose vitamin A supplements which can exacerbate intracranial hypertension.
\item Mass effect risk: Hold lumbar puncture until neuroimaging confirms absence of obstructive hydrocephalus or space-occupying lesion to prevent herniation.
\item Hemodynamic: Obtain immediate blood pressure measurement to rule out hypertensive emergency.
\end{itemize}
\item \textbf{Follow Up Or Escalation:} Return to ED immediately for sudden vision loss, severe refractory headache, projectile vomiting, or focal neurological deficits. Maintain inpatient admission or ED observation until Neurosurgery and Ophthalmology complete urgent evaluation, baseline labs (including hCG and BMP) are reviewed, and definitive imaging (Head\_MRV) is obtained to rule out cerebral venous sinus thrombosis.
\end{itemize}
\par\noindent\textcolor{AppendixRule}{\rule{\linewidth}{0.25pt}}\par
\paragraph{Recorded outcome.}
Gold diagnosis: \textbf{Idiopathic intracranial hypertension}. The no-MedSAM prediction was scored \textbf{incorrect}.

\subsection{Remote condition: Measurement learning with MedSAM}
\paragraph{Retrieved memory context.}
\begin{itemize}\setlength{\itemsep}{1pt}\setlength{\topsep}{2pt}
\item Acuity-Gated Cerebrovascular \& Headache Risk Stratification with Baseline-First Fallback and Prophylaxis-Bundle Verification
\item Adaptive Concurrent Baseline \& Empiric-Antimicrobial Neurological Workup Protocol
\end{itemize}
\paragraph{Complete observable interaction.}
\par\noindent
\noindent\colorbox{AppendixEnvironment}{\strut\textcolor{white}{\scriptsize\bfseries T1 ENVIRONMENT}}\quad
\begin{itemize}\setlength{\itemsep}{1pt}\setlength{\topsep}{2pt}
\item \textbf{Physical Examination:} \begin{itemize}\setlength{\itemsep}{1pt}\setlength{\topsep}{2pt}
\item \textbf{Focused Examination:} Bilateral optic disk swelling and retinal hemorrhages noted.
\end{itemize}
\item \textbf{Available Test Names:} \begin{itemize}\setlength{\itemsep}{1pt}\setlength{\topsep}{2pt}
\item NEJM\_Medical\_Image
\item Head\_MRI
\item Head\_MRV
\end{itemize}
\item \textbf{Required Test Names Before Final:} \begin{itemize}\setlength{\itemsep}{1pt}\setlength{\topsep}{2pt}
\item NEJM\_Medical\_Image
\end{itemize}
\item \textbf{Completion Instruction:} Request each required test with REQUEST TEST: <exact name> before returning DIAGNOSIS READY. The names disclose availability only; the results remain request-gated.
\end{itemize}
\par\noindent\textcolor{AppendixRule}{\rule{\linewidth}{0.25pt}}\par
\par\noindent
\noindent\colorbox{AppendixDoctor}{\strut\textcolor{white}{\scriptsize\bfseries T2 DOCTOR}}\quad
Please state medication and OTC use and allergies; alcohol amount/frequency; tobacco pack-years or vaping; recreational drugs; recent travel/exposures; relevant family history; and, if applicable, last menstrual period or pregnancy status.
\par\noindent\textcolor{AppendixRule}{\rule{\linewidth}{0.25pt}}\par
\par\noindent
\noindent\colorbox{AppendixPatient}{\strut\textcolor{white}{\scriptsize\bfseries T3 PATIENT}}\quad
I don't take any medications or have any allergies. I don't know about my family history, recent travel, or my last menstrual period and pregnancy status. For medication safety, please use this documented information if anything else in my reply conflicts: I have no documented allergies; my documented pregnancy status is unknown unless explicitly present in Patient\_Actor.
\par\noindent\textcolor{AppendixRule}{\rule{\linewidth}{0.25pt}}\par
\par\noindent
\noindent\colorbox{AppendixDoctor}{\strut\textcolor{white}{\scriptsize\bfseries T4 DOCTOR}}\quad
Could you describe any headaches you are experiencing, particularly their onset, severity, and whether they worsen when lying down or with coughing? Have you noticed any sudden vision changes, double vision, or nausea?
\par\noindent\textcolor{AppendixRule}{\rule{\linewidth}{0.25pt}}\par
\par\noindent
\noindent\colorbox{AppendixPatient}{\strut\textcolor{white}{\scriptsize\bfseries T5 PATIENT}}\quad
My headaches have been going on for about a week and feel quite severe, particularly when I lie down or bend over. I’ve also experienced brief moments where my vision suddenly grays out or blacks out, alongside persistent blurriness. I haven’t noticed any double vision or nausea.
\par\noindent\textcolor{AppendixRule}{\rule{\linewidth}{0.25pt}}\par
\par\noindent
\noindent\colorbox{AppendixDoctor}{\strut\textcolor{white}{\scriptsize\bfseries T6 DOCTOR}}\quad
REQUEST TEST: Head\_MRI
\par\noindent\textcolor{AppendixRule}{\rule{\linewidth}{0.25pt}}\par
\par\noindent
\noindent\colorbox{AppendixMeasurement}{\strut\textcolor{white}{\scriptsize\bfseries T7 MEASUREMENT}}\quad
RESULTS: Flattened posterior globes and an empty sella observed.
\par\noindent\textcolor{AppendixRule}{\rule{\linewidth}{0.25pt}}\par
\par\noindent
\noindent\colorbox{AppendixDoctor}{\strut\textcolor{white}{\scriptsize\bfseries T8 DOCTOR}}\quad
REQUEST TEST: Head\_MRV
\par\noindent\textcolor{AppendixRule}{\rule{\linewidth}{0.25pt}}\par
\par\noindent
\noindent\colorbox{AppendixMeasurement}{\strut\textcolor{white}{\scriptsize\bfseries T9 MEASUREMENT}}\quad
RESULTS: Transverse sinus stenoses present without evidence of obstruction or thrombosis.
\par\noindent\textcolor{AppendixRule}{\rule{\linewidth}{0.25pt}}\par
\par\noindent
\noindent\colorbox{AppendixDoctor}{\strut\textcolor{white}{\scriptsize\bfseries T10 DOCTOR}}\quad
Could you please specify your alcohol consumption frequency and amount, any tobacco use or vaping history, and whether you take any recreational substances? Additionally, what is your current height and weight?
\par\noindent\textcolor{AppendixRule}{\rule{\linewidth}{0.25pt}}\par
\par\noindent
\noindent\colorbox{AppendixPatient}{\strut\textcolor{white}{\scriptsize\bfseries T11 PATIENT}}\quad
I don't drink alcohol, use tobacco or vaping products, or take any recreational substances. I don't have my exact height and weight recorded, but I do know that I have severe obesity.
\par\noindent\textcolor{AppendixRule}{\rule{\linewidth}{0.25pt}}\par
\par\noindent
\noindent\colorbox{AppendixDoctor}{\strut\textcolor{white}{\scriptsize\bfseries T12 DOCTOR}}\quad
Could you please confirm your current pregnancy status and provide your exact height and weight to calculate your BMI?
\par\noindent\textcolor{AppendixRule}{\rule{\linewidth}{0.25pt}}\par
\par\noindent
\noindent\colorbox{AppendixPatient}{\strut\textcolor{white}{\scriptsize\bfseries T13 PATIENT}}\quad
I don't know my current pregnancy status, and I don't have my exact height and weight recorded. I can only tell you that I have severe obesity. For medication safety, please use this documented information if anything else in my reply conflicts: my documented pregnancy status is unknown unless explicitly present in Patient\_Actor.
\par\noindent\textcolor{AppendixRule}{\rule{\linewidth}{0.25pt}}\par
\par\noindent
\noindent\colorbox{AppendixDoctor}{\strut\textcolor{white}{\scriptsize\bfseries T14 DOCTOR}}\quad
REQUEST TEST: NEJM\_Medical\_Image
\par\noindent\textcolor{AppendixRule}{\rule{\linewidth}{0.25pt}}\par
\par\noindent
\noindent\colorbox{AppendixMeasurement}{\strut\textcolor{white}{\scriptsize\bfseries T15 MEASUREMENT}}\quad
\nopagebreak[4]\par\noindent
\fbox{\begin{minipage}{0.97\linewidth}
\textcolor{AppendixMeasurement}{\bfseries Image input to the Measurement Agent}\par
\noindent\begin{minipage}[t]{0.32\linewidth}
\vspace{0pt}
\centering\includegraphics[width=\linewidth]{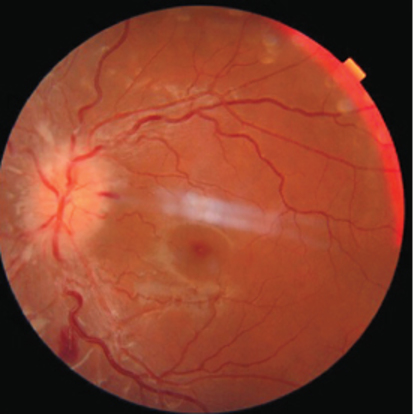}\\[-1mm]
\scriptsize Panel 0: fundus photograph
\end{minipage}\hfill
\begin{minipage}[t]{0.32\linewidth}
\vspace{0pt}
\centering\includegraphics[width=\linewidth]{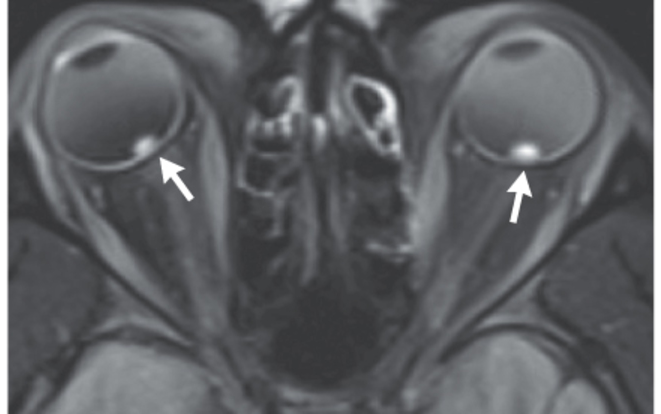}\\[-1mm]
\scriptsize Panel 1: orbital MRI
\end{minipage}\hfill
\begin{minipage}[t]{0.32\linewidth}
\vspace{0pt}
\centering\includegraphics[width=\linewidth]{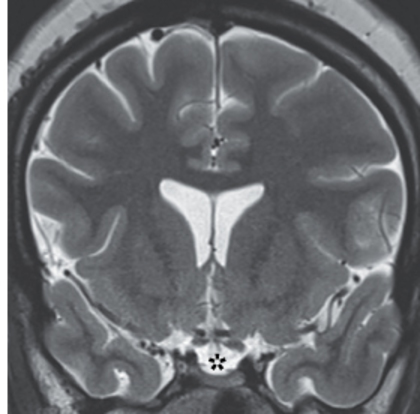}\\[-1mm]
\scriptsize Panel 2: coronal head MRI
\end{minipage}
\par\noindent\textcolor{AppendixMeasurement}{\bfseries MedSAM segmentation overlays}\par
\noindent\begin{minipage}[t]{0.32\linewidth}
\vspace{0pt}
\centering\includegraphics[width=\linewidth]{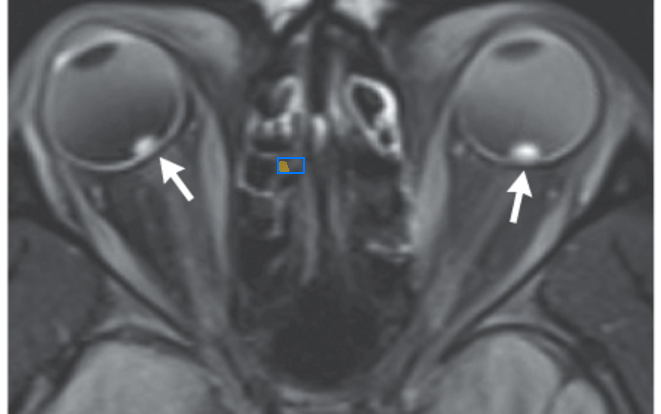}\\[-1mm]
\scriptsize Panel 1, ROI 0
\end{minipage}\hfill
\begin{minipage}[t]{0.32\linewidth}
\vspace{0pt}
\centering\includegraphics[width=\linewidth]{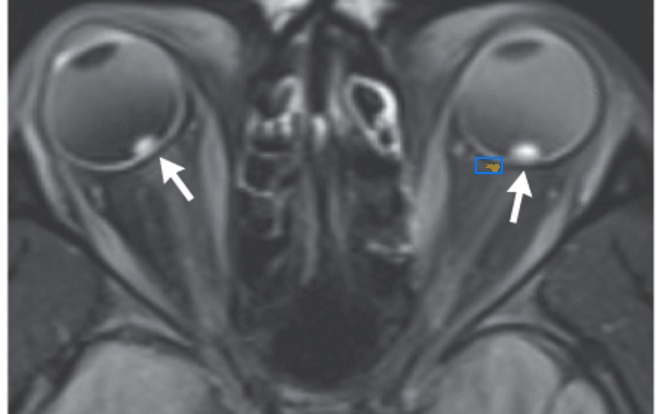}\\[-1mm]
\scriptsize Panel 1, ROI 1
\end{minipage}\hfill
\begin{minipage}[t]{0.32\linewidth}
\vspace{0pt}
\centering\includegraphics[width=\linewidth]{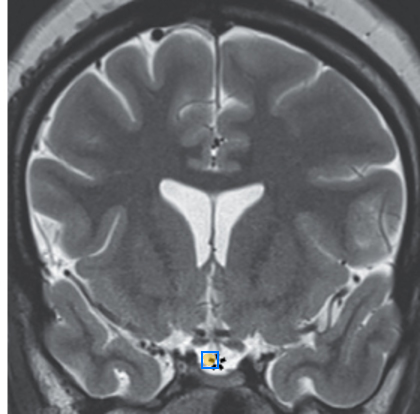}\\[-1mm]
\scriptsize Panel 2, ROI 0
\end{minipage}
\end{minipage}}
\par\noindent
\begin{itemize}\setlength{\itemsep}{1pt}\setlength{\topsep}{2pt}
\item \textbf{Report:} \begin{itemize}\setlength{\itemsep}{1pt}\setlength{\topsep}{2pt}
\item \textbf{Task Focus:} diagnosis
\item \textbf{Panel Findings:} \begin{itemize}\setlength{\itemsep}{1pt}\setlength{\topsep}{2pt}
\item panel index: 0; modality: Fundus Photography; visible findings: Color fundus photograph of the retina. The optic disc is visible on the left with distinct margins. Retinal vessels radiate from the disc. A bright, linear reflection artifact is present centrally over the macula. There are no obvious large hemorrhages or exudates in this view.; confidence: 0.95; roi boxes:
\item panel index: 1; modality: MRI (Orbits); visible findings: Axial MRI scan of the orbits showing both globes. White arrows point to hyperintense foci located at the posterior aspect of the globe, adjacent to the optic nerve head region bilaterally. These appear as small, rounded areas of high signal intensity within the vitreous cavity near the retinal surface.; confidence: 0.98; roi boxes: 420; 380; 460; 420; 720; 380; 760; 420
\item panel index: 2; modality: MRI (Brain); visible findings: Coronal T2-weighted MRI of the brain. An asterisk marks a focal area of abnormality at the inferior aspect of the midline, likely involving the cerebellar vermis or fourth ventricle region. The lesion appears hypointense relative to the surrounding CSF and brain parenchyma.; confidence: 0.95; roi boxes: 480; 850; 520; 890
\end{itemize}
\item \textbf{Mask Derived Observations:} \begin{itemize}\setlength{\itemsep}{1pt}\setlength{\topsep}{2pt}
\item[] \emph{Implementation note:} solidity divides integer selected-pixel area by OpenCV's continuous contour-hull area. For very small masks these discrete/continuous conventions can produce values slightly above one; solidity is an auxiliary report field and is not used for case selection, diagnosis, or scoring.
\item panel index: 1; modality: MRI (Orbits); box xyxy: 277; 157; 304; 173; quantitative features: valid: True; area ratio: 0.000402; prompt box coverage: 0.25463; component count: 1; largest component fraction: 1.0; centroid normalized xy: 0.4285; 0.4038; bbox normalized xyxy: 0.4221; 0.3889; 0.4387; 0.4203; bbox extent: 0.769231; circularity: 0.921625; solidity: 1.128205; inside gray mean: 80.464; surrounding ring gray mean: 75.806; inside minus ring gray: 4.658; edge fraction inside: 0.0; touches image boundary: False; touches prompt boundary: True
\item panel index: 1; modality: MRI (Orbits); box xyxy: 475; 157; 502; 173; quantitative features: valid: True; area ratio: 0.00027; prompt box coverage: 0.171296; component count: 1; largest component fraction: 1.0; centroid normalized xy: 0.746; 0.403; bbox normalized xyxy: 0.7352; 0.3961; 0.7564; 0.4155; bbox extent: 0.660714; circularity: 0.570172; solidity: 1.049645; inside gray mean: 82.041; surrounding ring gray mean: 83.578; inside minus ring gray: -1.537; edge fraction inside: 0.0; touches image boundary: False; touches prompt boundary: True
\item panel index: 2; modality: MRI (Brain); box xyxy: 201; 351; 218; 368; quantitative features: valid: True; area ratio: 0.000656; prompt box coverage: 0.394464; component count: 1; largest component fraction: 1.0; centroid normalized xy: 0.4967; 0.8671; bbox normalized xyxy: 0.4833; 0.8551; 0.5119; 0.8841; bbox extent: 0.791667; circularity: 0.984702; solidity: 1.123153; inside gray mean: 183.833; surrounding ring gray mean: 181.55; inside minus ring gray: 2.283; edge fraction inside: 0.192982; touches image boundary: False; touches prompt boundary: True
\end{itemize}
\item \textbf{Cross Panel Synthesis:} The fundus photography shows a relatively normal-appearing retina with a central reflection artifact, lacking obvious large hemorrhages or exudates despite the clinical note of swelling. The orbital MRI reveals bilateral, symmetric, hyperintense foci at the posterior pole of the globes, corresponding anatomically to the optic nerve head region. The brain MRI demonstrates a focal hypointense lesion in the midline inferiorly, potentially involving the cerebellar vermis or fourth ventricle. The combination of bilateral posterior eye findings and a midline brain lesion suggests a systemic or syndromic process affecting these specific anatomical regions.
\item \textbf{Limitations:} The segmentation masks for the orbital MRI lesions show high circularity but low prompt box coverage, indicating they capture the core of the hyperintense foci but may not fully delineate their irregular boundaries against the vitreous. The brain lesion mask has a higher coverage ratio but its exact anatomical relationship to the fourth ventricle vs. vermis is difficult to confirm without sagittal/coronal multi-planar reconstruction. The fundus image lacks color contrast resolution to definitively rule out subtle micro-hemorrhages or exudates mentioned in the clinical history.
\item \textbf{Segmentation Assessment:} Masks serve only as approximate anatomical localization aids, not precise pathological quantifiers. Cross-verification against raw pixels confirms the orbital masks align with the hyperintense foci pointed out by arrows. The brain mask aligns with the asterisk-marked region. However, automated masks cannot inherently differentiate subtle pathological textures from normal background tissue; final morphological assessments rely on direct visual inspection of raw images.
\end{itemize}
\item \textbf{Medsam Enabled:} True
\item \textbf{Task Focus:} \begin{itemize}\setlength{\itemsep}{1pt}\setlength{\topsep}{2pt}
\item \textbf{Answer Type:} diagnosis
\item \textbf{Measurement Guidance:} The requested answer type is a diagnosis; provide discriminative visible morphology but leave disease naming to the Doctor.
\end{itemize}
\item \textbf{Provided Exam Context:} \begin{itemize}\setlength{\itemsep}{1pt}\setlength{\topsep}{2pt}
\item \textbf{Focused Examination:} Bilateral optic disk swelling and retinal hemorrhages noted.
\end{itemize}
\item \textbf{Measurement Skill Ids:} \begin{itemize}\setlength{\itemsep}{1pt}\setlength{\topsep}{2pt}
\item vm\_evolved\_g38\_48bfeb9bde45
\item vm\_evolved\_g22\_71611f3e24ce
\item vm\_evolved\_g225\_2b3401c40f64
\end{itemize}
\item \textbf{Segmentation Results:} \begin{itemize}\setlength{\itemsep}{1pt}\setlength{\topsep}{2pt}
\item \begin{itemize}\setlength{\itemsep}{1pt}\setlength{\topsep}{2pt}
\item \textbf{Panel Index:} 1
\item \textbf{Modality:} MRI (Orbits)
\item \textbf{Box Xyxy:} 277; 157; 304; 173
\item \textbf{Mask Area Ratio:} 0.000402
\item \textbf{Mask Nonempty:} True
\item \textbf{Quantitative Features:} valid: True; area ratio: 0.000402; prompt box coverage: 0.25463; component count: 1; largest component fraction: 1.0; centroid normalized xy: 0.4285; 0.4038; bbox normalized xyxy: 0.4221; 0.3889; 0.4387; 0.4203; bbox extent: 0.769231; circularity: 0.921625; solidity: 1.128205; inside gray mean: 80.464; surrounding ring gray mean: 75.806; inside minus ring gray: 4.658; edge fraction inside: 0.0; touches image boundary: False; touches prompt boundary: True
\end{itemize}
\item \begin{itemize}\setlength{\itemsep}{1pt}\setlength{\topsep}{2pt}
\item \textbf{Panel Index:} 1
\item \textbf{Modality:} MRI (Orbits)
\item \textbf{Box Xyxy:} 475; 157; 502; 173
\item \textbf{Mask Area Ratio:} 0.00027
\item \textbf{Mask Nonempty:} True
\item \textbf{Quantitative Features:} valid: True; area ratio: 0.00027; prompt box coverage: 0.171296; component count: 1; largest component fraction: 1.0; centroid normalized xy: 0.746; 0.403; bbox normalized xyxy: 0.7352; 0.3961; 0.7564; 0.4155; bbox extent: 0.660714; circularity: 0.570172; solidity: 1.049645; inside gray mean: 82.041; surrounding ring gray mean: 83.578; inside minus ring gray: -1.537; edge fraction inside: 0.0; touches image boundary: False; touches prompt boundary: True
\end{itemize}
\item \begin{itemize}\setlength{\itemsep}{1pt}\setlength{\topsep}{2pt}
\item \textbf{Panel Index:} 2
\item \textbf{Modality:} MRI (Brain)
\item \textbf{Box Xyxy:} 201; 351; 218; 368
\item \textbf{Mask Area Ratio:} 0.000656
\item \textbf{Mask Nonempty:} True
\item \textbf{Quantitative Features:} valid: True; area ratio: 0.000656; prompt box coverage: 0.394464; component count: 1; largest component fraction: 1.0; centroid normalized xy: 0.4967; 0.8671; bbox normalized xyxy: 0.4833; 0.8551; 0.5119; 0.8841; bbox extent: 0.791667; circularity: 0.984702; solidity: 1.123153; inside gray mean: 183.833; surrounding ring gray mean: 181.55; inside minus ring gray: 2.283; edge fraction inside: 0.192982; touches image boundary: False; touches prompt boundary: True
\end{itemize}
\end{itemize}
\item \textbf{Limitations:} MedSAM masks are prompt-conditioned localization aids; preserve the raw-image findings.
\end{itemize}
\par\noindent\textcolor{AppendixRule}{\rule{\linewidth}{0.25pt}}\par
\par\noindent
\noindent\colorbox{AppendixDoctor}{\strut\textcolor{white}{\scriptsize\bfseries T16 DOCTOR}}\quad
\begin{itemize}\setlength{\itemsep}{1pt}\setlength{\topsep}{2pt}
\item \textbf{Diagnosis:} Idiopathic Intracranial Hypertension (IIH)
\item \textbf{Differential Diagnoses:} \begin{itemize}\setlength{\itemsep}{1pt}\setlength{\topsep}{2pt}
\item Posterior Fossa Mass Lesion with Obstructive Hydrocephalus
\item Cerebral Venous Sinus Thrombosis
\item Meningitis/Infectious Etiology
\item Malignant Hypertension
\end{itemize}
\item \textbf{Key Evidence:} \begin{itemize}\setlength{\itemsep}{1pt}\setlength{\topsep}{2pt}
\item Severe obesity
\item Headaches worsening with recumbency and bending over (positional/Valsalva exacerbation)
\item Transient visual obscurations and persistent blurriness
\item Initial focused examination showing bilateral optic disk swelling and retinal hemorrhages
\item MRI demonstrating flattened posterior globes, empty sella, and transverse sinus stenoses without thrombosis
\item MRI revealing a focal hypointense midline inferior lesion near the cerebellar vermis/fourth ventricle
\end{itemize}
\item \textbf{Tests Used:} \begin{itemize}\setlength{\itemsep}{1pt}\setlength{\topsep}{2pt}
\item Head\_MRI
\item Head\_MRV
\item NEJM\_Medical\_Image
\end{itemize}
\item \textbf{Treatment Prescription Plan:} \begin{itemize}\setlength{\itemsep}{1pt}\setlength{\topsep}{2pt}
\item Urgent Neurosurgery consultation and recommendation for hospital admission for continuous neurologic monitoring and evaluation of the posterior fossa lesion and elevated ICP.
\item Ophthalmology referral for formal visual field testing and standardized fundus documentation to reconcile initial exam findings with imaging.
\item DO NOT START UNTIL VERIFIED: Acetazolamide 500 mg orally twice daily (titrating to 1000 mg twice daily) to reduce CSF production; withhold pending confirmed non-pregnancy status and baseline BMP/LFTs.
\item Non-drug management: Head elevation to 30 degrees, structured weight reduction program targeting 5-10\% body weight loss, dietary sodium restriction (<2g/day), strict avoidance of Vitamin A supplements and tetracycline-class antibiotics.
\item Obtain mandatory baseline laboratories immediately: CBC, comprehensive metabolic panel (BMP), coagulation profile, liver function tests, and quantitative serum/urine pregnancy test prior to any pharmacotherapy.
\item Strict avoidance of NSAIDs, anticoagulants, and antiplatelet agents until hemorrhage risk and mass effect are fully characterized.
\end{itemize}
\item \textbf{Safety Checks:} \begin{itemize}\setlength{\itemsep}{1pt}\setlength{\topsep}{2pt}
\item Pregnancy status is currently unknown; acetazolamide carries teratogenic risk and must be withheld until a quantitative serum/urine pregnancy test confirms non-pregnancy.
\item Renal and hepatic baselines are unverified; acetazolamide requires renal clearance and may precipitate metabolic acidosis or electrolyte depletion. Baseline BMP/LFTs are mandatory before first dose.
\item No known drug allergies documented; explicit verification required prior to dispensing.
\item Posterior fossa lesion and signs of elevated ICP require urgent neurosurgical evaluation to rule out obstructive hydrocephalus or herniation risk.
\end{itemize}
\item \textbf{Follow Up Or Escalation:} Admit to a monitored neurological unit for frequent neuro checks and urgent specialist consultations; return to Emergency Department immediately for sudden permanent vision loss, thunderclap headache, projectile vomiting, or altered mental status.
\end{itemize}
\par\noindent\textcolor{AppendixRule}{\rule{\linewidth}{0.25pt}}\par
\paragraph{Recorded outcome.}
Gold diagnosis: \textbf{Idiopathic intracranial hypertension}. The MedSAM prediction was scored \textbf{correct} by the automatic diagnosis metric; this label does not endorse the visual evidence or management plan.

\paragraph{Interpretation.}
This pair is a qualitative failure-boundary illustration rather than positive clinical evidence or a causal estimate. In both paths, source annotations are overinterpreted as pathology; in the MedSAM path, an unverified posterior-fossa finding propagates into the differential and escalation plan despite the correct final label. The diagnosis-blind release checks enforce provenance, prerequisites, and conservative disposition but cannot determine whether an image interpretation is clinically true. This trace therefore exposes a limitation of the current governance stack and motivates independent image adjudication; it must not be read as evidence that MedSAM improved this case.
\endgroup

\fi

\end{document}